\documentclass[twocolumn]{article}
\usepackage{arxiv}
\usepackage{cite}
\usepackage[pdftex]{graphicx}
\graphicspath{{}}
\usepackage{amsmath}
\usepackage[explicit]{titlesec}
\titleformat{\paragraph}[runin]
  {\normalfont\normalsize\bfseries}{}{15pt}{\theparagraph#1.}
  \titleformat{name=\paragraph,numberless}[runin]
  {\normalfont\normalsize\bfseries}{}{15pt}{#1.}
\titlespacing*{\paragraph}{0pt}{0ex plus 0ex minus 0ex}{15pt}

\usepackage{placeins}
\usepackage{caption}

\usepackage[font={normal}]{caption} 
\usepackage[font={small}]{subcaption} 

\usepackage{amssymb} 

\usepackage{algpseudocode} 
\usepackage{subfiles} 
\usepackage{mathrsfs} 
\usepackage{comment}  
\newcommand{\argmin}{\operatornamewithlimits{argmin}}
\newenvironment{Itemize}%
{
\setlength{\leftmargini}{9pt}%
\begin{itemize}%
\setlength{\itemsep}{0pt}%
\setlength{\topsep}{0pt}%
\setlength{\partopsep}{0pt}%
\setlength{\parskip}{0pt}}%
{\end{itemize}}

\usepackage[utf8]{inputenc}
\usepackage[english]{babel}
\usepackage{float}
\usepackage{bbm}
\usepackage{amsfonts}
\usepackage{multirow}
\usepackage{array}
\usepackage{arydshln}
\usepackage{booktabs}
\usepackage{xcolor}
\definecolor{DarkGreen}{rgb}{0.43, 0.68, 0.28}
\usepackage{tabularx}

\usepackage{blindtext}

\usepackage{autobreak}
\usepackage{amsmath}
\usepackage{enumitem}
\setlist[itemize]{noitemsep,topsep=3pt,leftmargin=*}

\usepackage[original]{abstract} %
\usepackage{balance}

\usepackage[pagebackref,breaklinks,colorlinks]{hyperref}

\newcommand{\name} {TokenCut}
\usepackage[capitalize]{cleveref}
\crefname{section}{Sec.}{Secs.}
\Crefname{section}{Section}{Sections}
\Crefname{table}{Table}{Tables}
\crefname{table}{Tab.}{Tabs.}

\usepackage{colortbl}
\usepackage{pifont}%
\usepackage{makecell}

\definecolor{LightGray}{rgb}{0.9,0.9,0.9}

\definecolor{cssgreen}{rgb}{0.0, 0.5, 0.0}
\definecolor{cssred}{rgb}{1, 0, 0.0}
\newcommand{\feat}{\mathbf{v}}
\newcommand{\indicator}{\mathbf{x}}
\newcommand{\indicatory}{\mathbf{y}}
\newcommand{\indicatorz}{\mathbf{z}}

\newcommand{\degree}{\mathbf{d}}

\newcommand{\nbnode}{N}

\newcommand{\graph}{\mathcal{G}}
\newcommand{\node}{\mathcal{V}}
\newcommand{\edge}{\mathcal{E}}
\newcommand{\Degree}{\mathbf{D}}
\newcommand{\J}{\mathcal{J}}

\newcommand{\connect}{C}

\newcommand{\simi}{S}
\newcommand{\minsimi}{\tau}

\newcommand{\cmark}{\ding{51}}%

\newcommand{\lmm}[1]{\textcolor{red}{#1}}

\newcommand{\CUT}[1]{}

\hypersetup{
  colorlinks=true,
  citecolor=cssgreen,
  urlcolor=red,
  pdftitle={TokenCut: Segmenting Objects in Images and Videos with Self-supervised Transformer and Normalized Cut},
  pdfauthor={Yangtao Wang, Xi Shen, Yuan Yuan, Yuming Du, Maomao Li, Shell Xu Hu, James L. Crowley, Dominique Vaufreydaz},
  pdfkeywords={Unsupervised object discovery, unsupervised saliency detection, unsupervised video object segmentation, self-supervised transformer, normalized cut},
}

\title{\name: Segmenting Objects in Images and Videos with Self-supervised Transformer and Normalized Cut}

\author{Yangtao Wang$^{1*}$, Xi Shen$^{2*}$, Yuan Yuan$^3$,  Yuming Du$^4$,  Maomao Li$^2$, Shell Xu Hu$^5$ \\ \textbf{James L. Crowley}$^1$, \textbf{\href{https://research.vaufreydaz.org/}{\color{black}Dominique Vaufreydaz}}$^1$ \\
$^1$ Univ. Grenoble Alpes, CNRS, Grenoble INP, LIG, 38000 Grenoble, France\\
$^2$ Tencent AI Lab \hspace{5mm}
$^3$ MIT CSAIL 
$^4$ LIGM (UMR 8049) - Ecole des Ponts, UPE \\
$^5$ Samsung AI Center, Cambridge \hspace{5mm}
}

\begin{document}

\twocolumn[{%
  \begin{@twocolumnfalse}
    \maketitle
  \end{@twocolumnfalse}
}]

\setcounter{footnote}{0}
\renewcommand{\thefootnote}{*}
\footnotetext[1]{Corresponding Author}

\begin{abstract}
In this paper, we describe a graph-based algorithm that uses the
features obtained by a self-supervised transformer to detect and segment salient objects in images and videos. 
With this approach, the image patches that compose an image or video are organised into a fully connected graph, in which the edge between each pair of patches is labeled with a similarity score based on the features learned by the transformer. Detection and segmentation of salient objects can then be formulated as a graph-cut problem and solved using the classical Normalized Cut algorithm. Despite the simplicity of this approach, it achieves state-of-the-art results on several common image and video detection and segmentation tasks. For unsupervised object discovery, this approach outperforms the competing approaches by a margin of 6.1\%, 5.7\%, and 2.6\% when tested with the VOC07, VOC12, and COCO20K datasets. For the unsupervised saliency detection task in images, this method improves the score for Intersection over Union (IoU) by 4.4\%, 5.6\% and 5.2\%. When tested with the ECSSD, DUTS, and DUT-OMRON datasets. This method also achieves competitive results for unsupervised video object segmentation tasks with the DAVIS, SegTV2, and FBMS datasets. Our implementation is available at  \href{https://www.m-psi.fr/Papers/TokenCut2022/}{https://www.m-psi.fr/Papers/TokenCut2022/}.

\end{abstract}

\section{Introduction}

Detecting and segmenting salient objects in an image or video are fundamental problems in computer vision with applications in real-world vision systems for robotics, autonomous driving, traffic monitoring, manufacturing, and embodied artificial intelligence~\cite{wu2017squeezedet,geiger2013vision,xia2018dota}. However, current approaches rely on supervised learning requiring large data sets of high-quality, annotated training data~\cite{lin2014microsoft}. The high cost of this approach becomes even more apparent when using transfer learning to adapt a pre-trained object detector to a new application domain. Researchers have attempted to overcome this barrier using active learning~\cite{aghdam2019active, siddiqui2020viewal}, semi-supervised learning~\cite{liu2021unbiased, chen2021semi}, and weakly-supervised learning~\cite{ren2020instance, ke2021universal} with limited results.
In this paper, we report on results of an effort to use features provided by transformers trained with self-supervised learning, obviating the need for expensive annotated training data. 

\begin{figure}[!t]
	\centering
	 \begin{subfigure}[b]{\linewidth}
         \centering
         \includegraphics[width=0.8\linewidth, height=2.6cm]{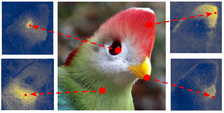}
         \caption{Attention maps associated to different patches}
         \label{fig:teaser_a}
     \end{subfigure}
     \hfill
     \begin{subfigure}[b]{\linewidth}
         \centering
         \includegraphics[width=0.7\linewidth, height=4.0cm]{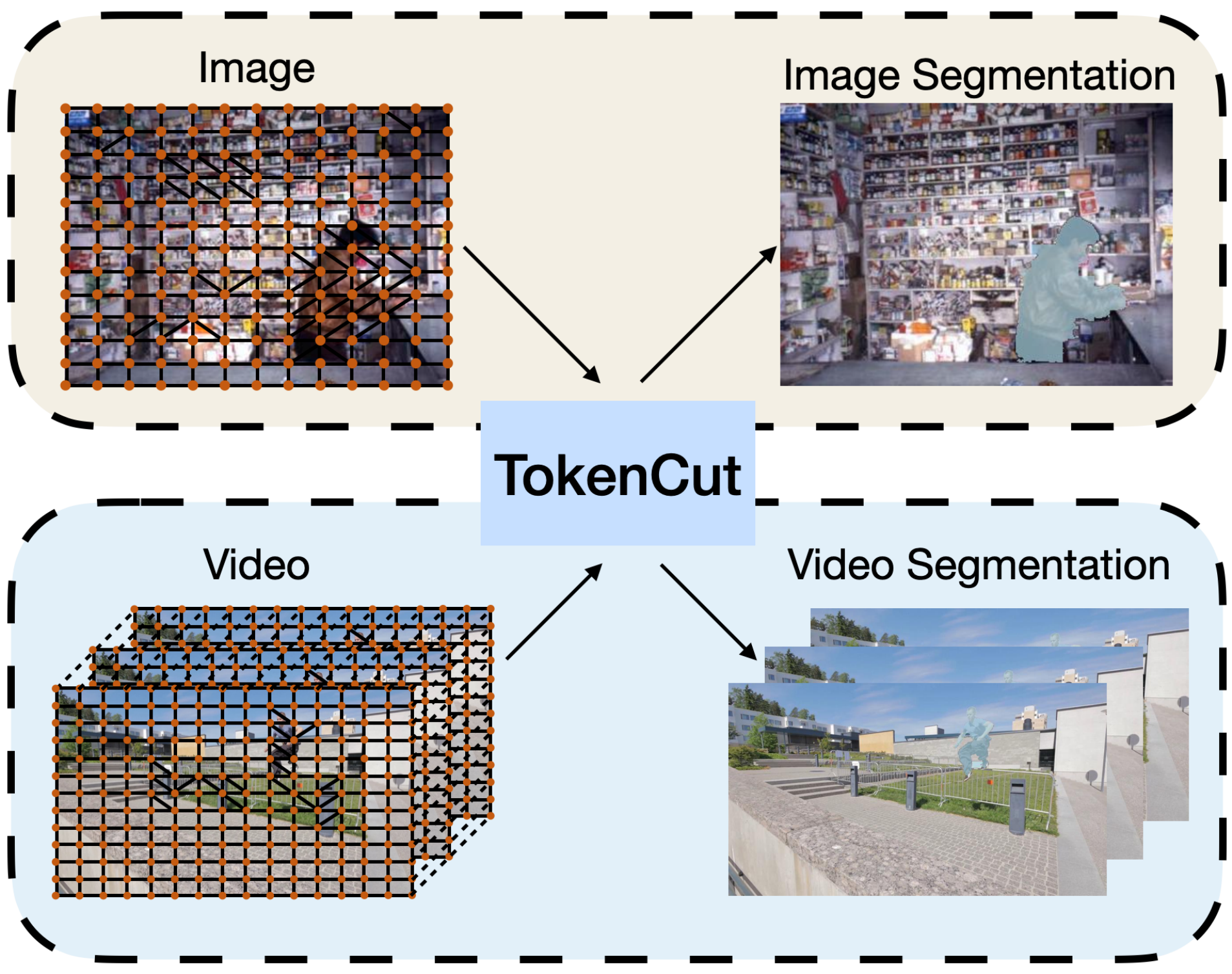}
         \caption{A unified method for image and video segmentation.}
         \label{fig:teaser}
     \end{subfigure}
    \vspace{-2mm} 
	\caption{Attention maps associated with different patches highlight different regions of the object (Fig.~\ref{fig:teaser_a}), which motivates us to build a unified graph-based solution for unsupervised image and video segmentation (Fig.~\ref{fig:teaser}).}
\end{figure}

Vision transformers trained with self-supervised learning ~\cite{dosovitskiy2020image,caron2021emerging}, such as DINO~\cite{caron2021emerging} and MAE~\cite{he2022masked, feichtenhofer2022masked, tong2022videomae} have been shown to outperform supervised training on downstream tasks. 
In particular, the attention maps associated with patches typically contain meaningful semantic information (Fig.~\ref{fig:teaser_a}). 
For example, experiments with DINO~\cite{caron2021emerging} indicate that the attention maps of the class token highlight salient object regions. 
However, such attention maps are noisy and cannot be directly used to detect or segment objects. 

The authors of LOST~\cite{simeoni2021localizing} have shown that the learned features from DINO can be used to build a graph and segment objects using the inverse degrees of nodes. Specifically, LOST employs a heuristic seed expansion strategy to accommodate noise and detect a single bounding box for a foreground object. We have investigated whether such learned features can be used with a graph-based approach to detect and segment salient objects in images and videos (Fig.~\ref{fig:teaser}), formulating the segmentation problem using the classic normalised cut algorithm (Ncut)~\cite{shi2000normalized}. 

\begin{figure*}[!thbp]
	\centering
	 \begin{subfigure}[b]{0.41\textwidth}
         \centering
         \includegraphics[width=\textwidth]{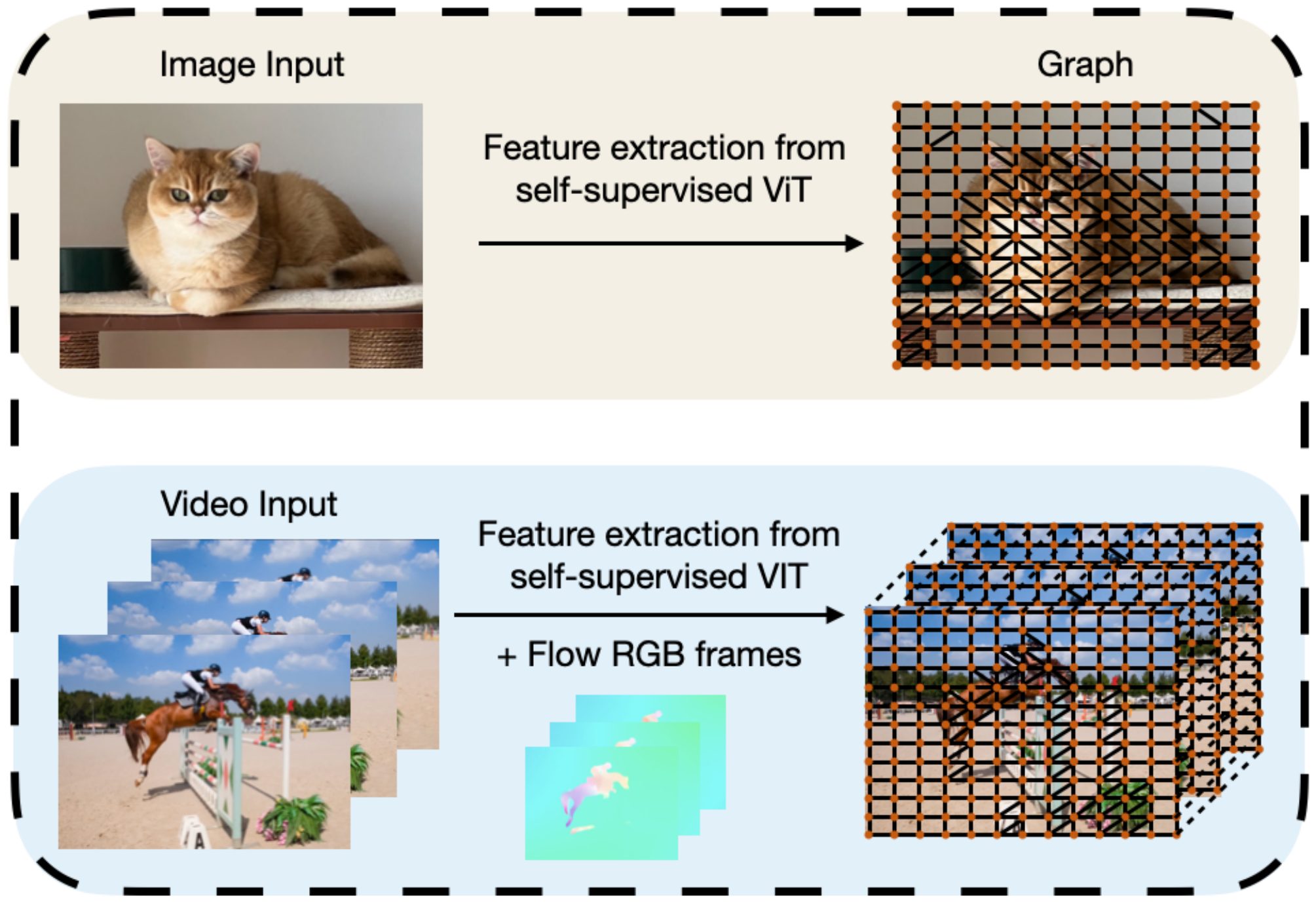}
         \caption{Graph Construction}
         \label{fig:graph_building}
     \end{subfigure}
     \hfill
     \begin{subfigure}[b]{0.38\textwidth}
         \centering
         \includegraphics[width=\textwidth]{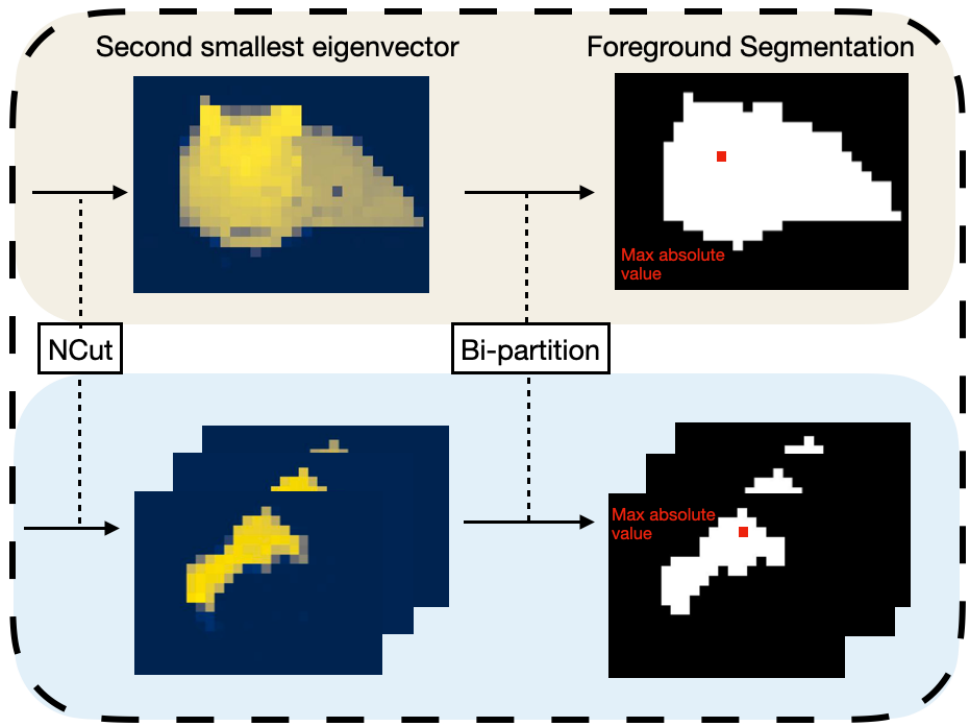}
         \caption{Graph Cut}
         \label{fig:graph_cut}
     \end{subfigure}
     \hfill
     \begin{subfigure}[b]{0.2\textwidth}
         \centering
         \includegraphics[width=\textwidth]{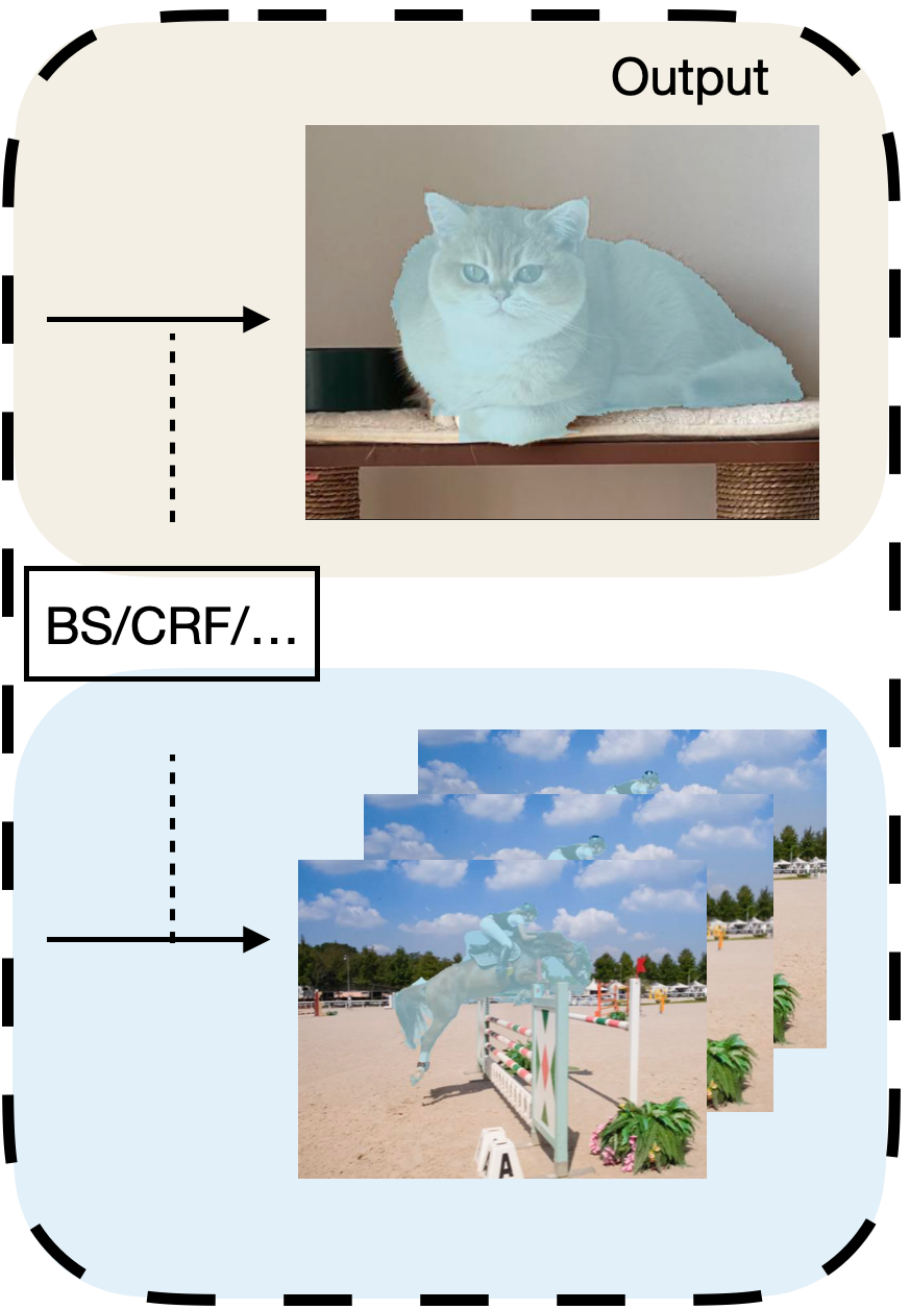}
         \caption{Edge Refinement}
         \label{fig:edge_refiner}
     \end{subfigure}
     
	\caption{An overview of the TokenCut approach. The algorithm constructs a fully connected graph in which the nodes are image patches and the edges are similarities between the image patches using transformer features. Object segmentation is then solved using the Ncut algorithm~\cite{shi2000normalized}. Bi-partition of the graph using the second smallest eigenvector allows to detect foreground object. A Bilateral Solver~\cite{barron2016fast} (BS) or Conditional Random Field~\cite{krahenbuhl2011efficient} (CRF) can be used for edge refinement.}
	\label{fig:method}	
\end{figure*}

In this paper we describe \name, a unified graph-based approach for image and video segmentation using features provided by self-supervised learning. The processing pipeline for this approach, illustrated in Fig.~\ref{fig:method}, is composed of three steps: \textit{1)} graph construction, \textit{2)} graph cut, \textit{3)} edge refinement. In the graph construction step, the algorithm uses image patches as nodes and uses features provided by self-supervised learning to describe the similarity between pairs of nodes.  For images, edges are labeled with a score for the similarity of patches based on learned features for RGB appearance.
For videos, edge labels combine similarities of learned features for RGB appearance and optical flow.

To cut the graph, we rely on the classic normalized cut (Ncut) algorithm
to group self-similar regions and delimit the salient objects. 
We solve the graph-cut problem using spectral clustering with generalized eigen-decomposition. The second smallest eigenvector provides a cutting solution indicating the likelihood that a token belongs to a foreground object, which allows us to design a simple post-processing step to obtain a foreground mask. %
We also show that standard algorithmns for edge-aware refinement, such as Conditional Random Field~\cite{krahenbuhl2011efficient} (CRF) and Bilateral Solver~\cite{barron2016fast} (BS) can be used to refine the masks for detailed object boundary detection. This approach can be considered as a run-time adaptation method, because the model can be used to process an image or video without the need to retrain the model.

Despite its simplicity, \name~significantly improves unsupervised saliency detection in images. Specifically, it achieves 77.7\%, 62.8\%. 61.9\% mIoU on the ECSSD~\cite{shi2015hierarchical}, DUTS~\cite{wang2017learning} and DUT-OMRON~\cite{yang2013saliency} respectively, and outperforms the previous state-of-the-art by a margin of 4.4\%, 5.6\% and 5.2\%. For unsupervised video segmentation, \name~achieves competitive results on DAVIS~\cite{perazzi2016benchmark}, FBMS~\cite{ochs2013segmentation}, SegTV2~\cite{li2013video}. Additionally, \name~also obtains important improvement on unsupervised object discovery. For example, \name~outperforms DSS~\cite{melaskyriazi2022deep}, which is a concurrent work, by a margin of 6.1\%, 5.7\%,
and 2.6\% respectively on the VOC07~\cite{pascal-voc-2007}, VOC12~\cite{pascal-voc-2012}, COCO20K~\cite{lin2014microsoft}.

In summary, the main contributions of this paper are as follows:
\begin{Itemize}

\item We describe \name~, a simple and unified approach to segment objects in images and videos that does not require human annotations for training.  \footnote{The implementation is available at \href{https://www.m-psi.fr/Papers/TokenCut2022/}{https://www.m-psi.fr/Papers/TokenCut2022/}. An online demo is accessible at \href{https://huggingface.co/spaces/yangtaowang/TokenCut}{https://huggingface.co/spaces/yangtaowang/TokenCut} (last access May 2023).}

\item We show that  \name~significantly outperforms previous state-of-the-art methods unsupervised saliency detection and unsupervised object discovery on images. As a training-free method, \name~achieves competitive performance on unsupervised video segmentation compared to the state-of-the-art methods.

\item We provide a detailed analysis on the \name~to validate the design of the proposed approach.

\end{Itemize}

\section{Related Work}
\paragraph*{Self-supervised vision transformers}

ViT~\cite{dosovitskiy2020image} has shown that the transformer architecture~\cite{vaswani2017attention} can be effective for computer vision tasks using supervised learning. Recently, many variants of ViT have been proposed to learn image encoders in a self-supervised manner. MoCo-V3~\cite{chen2021empirical} demonstrates that using contrastive learning on ViT can achieve strong results. DINO~\cite{caron2021emerging} shows that transformers can be trained with self-distillation loss~\cite{hinton2015distilling} and shows that the features learn by ViT contain explicit information useful for image semantic segmentation. Inspired by BERT, several approaches~\cite{devlin2018bert, li2021mst, bao2021beit, he2022masked} learn by missing token replacement, masking some tokens from the input input and learning to recover the missing tokens in the output.

\paragraph*{Unsupervised object discovery} 
Given a group of images, unsupervised object discovery seeks to discover and delimit similar objects that appear in multiple images.
Early research~\cite{joulin2010discriminative,joulin2012multi,vicente2011object,hsu2018co,chen2020show}  
formulated the problem using an hypothesis about the frequency of object occurrences. Other researchers formulated object detection as an optimization problem over bounding box proposals~\cite{tang2014co,cho2015unsupervised,vo2019unsupervised,vo2020toward} or as a ranking problem~\cite{vo2021large}.

Recently, LOST~\cite{simeoni2021localizing} significantly improved the state-of-the-art for unsupervised object discovery. LOST extracts features using a self-supervised transformer based on DINO~\cite{caron2021emerging} and designs a heuristic seed expansion strategy to obtain a single object region. 
A concurrent work DSS~\cite{melaskyriazi2022deep} designs a weighted graph over patches using self-supervised transformer features as well as a KNN based image matting algorithm using a color affinity matrix. The eigen-decomposition of the affinity matrix is computed to obtain a coarse object  mask. Following LOST~\cite{simeoni2021localizing}, DSS also assumes that the foreground object occupies a smaller region than the background. While DSS has a  similar eigen-attention map as~\name, DSS is less able to detect large objects.

As with LOST, \name~also uses  features obtained with self-supervised learning. However, rather than  relying on the attention map of some specific nodes, \name~forms a fully connected graph of image tokens, with edges labeled with a similarity score between tokens based on transformer features. The classical  Ncut~\cite{shi2000normalized} algorithm is then used to detect and segment image objects.%

\paragraph*{Unsupervised saliency detection}
Unsupervised saliency detection seeks to segment a salient object within an image. 
In this paper, we show that incorporating a simple post-processing step into \name~for unsupervised object discovery can provide a strong baseline method for unsupervised saliency detection. Earlier works on Unsupervised saliency detection~\cite{yan2013hierarchical, zhu2014saliency, jiang2013salient, li2015weighted} use techniques such as color contrast~\cite{cheng2014global}, background priors~\cite{wei2012geodesic}, or super-pixels~\cite{li2015weighted, yang2013saliency}. More recently, unsupervised deep models~\cite{ zhang2018deep, nguyen2019deepusps}  have been used for saliency detection using noisy pseudo-labels generated from different handcrafted saliency methods. ~\cite{voynov2021object} shows that unsupervised GANs can differentiate between foreground and background pixels and generate high-quality saliency masks. SelfMask~\cite{shin2022selfmask} use a spectral clustering method with self-supervised features to group pixels into a set of candidate clusters. SelfMask is trained by selecting the salient masks from the set of spectral clusters as pseudo-masks for supervision using cluster voting scheme.

\paragraph*{Unsupervised video segmentation}
Given an unlabeled video, unsupervised video segmentation aims to generate pixel-level masks for the object of interest in the video. Prior works segment objects by selecting super-pixels~\cite{koh2017primary}, learning flattened 3D object representations~\cite{lao2018extending}, constructing an adversarial network to mask a region such that the model can predict the optical flow of the masked region~\cite{yang2019unsupervised}, or reconstructing the optical flow in a self-supervised manner~\cite{yang2021self}, etc.
DyStaB~\cite{yang2021dystab} first partitions the motion field by minimizing the temporal consistent mutual information and then uses the segments to learn the object detector, in which the models are jointly trained with a bootstrapping strategy. The deformable sprites method (DeSprite)~\cite{ye2022deformable} trains a video auto-encoder to segment the object of interest by decomposing the video into layers of persistent motion groups. In contrast to these methods~\cite{yang2019unsupervised, yang2021self, yang2021dystab}, our proposed method does not require prior training on videos. Compared with methods~\cite{koh2017primary, lao2018extending} that do not train on videos, our method achieves superior performance.

\section{Approach: \name}
\label{sec:approach}

In this section, we present {\name}, a unified algorithm that can be used to segment salient objects in an image or moving objects in a video. Our approach, illustrated in Fig.~\ref{fig:method}, is based on a graph where the nodes are visual patches from either an image or a sequence of frames, and the edges are similarities between the features of the nodes based on the features provided by a visual transformer trained with  self-supervised learning.

This section is organised as follows: we first briefly review  vision transformers and the Normalized Cut algorithm
in Section~\ref{sec:visiontransformers} and Section~\ref{sec:ncut}. 
We then describe  the \name~algorithm for object detection and segmentation in images and videos in Section~\ref{sec:tokencut}.

\subsection{Background}
\subsubsection{Vision Transformers}
\label{sec:visiontransformers}
The Vision Transformer has been proposed in~\cite{dosovitskiy2020image}. The key idea is to process an image with transformer~\cite{vaswani2017attention} architectures using non-overlapping patches as tokens. For an image with size $H \times W$, a vision transformer takes non-overlapping $ K \times K $ image patches as inputs, resulting in $N=HW/K^{2}$ patches. 
Each patch is used as a token, described by a vector of numerical features that provide an embedding. An extra learnable token, denoted as a class token $CLS$, is used to represent the aggregated information of the entire set of patches. A positional encoding is added to $CLS$ token and the set of patch tokens, and the resulting vector is fed to a standard Vision Transformer with self-attention~\cite{vaswani2017attention} and layer normalization~\cite{ba2016layer}.

The Vision Transformer is composed of several stacked layers of encoders, each with feed-forward networks and multiple attention heads for self-attention, paralleled with skip connections. 
For the \name~algorithm, we use the Vision Transformer, trained with self-supervised learning. We  extract latent features from the final layer as the input features for \name.

\subsubsection{Normalized Cut (Ncut)}
\label{sec:ncut}
\paragraph*{Graph partitioning} 
Given a graph $\mathcal{G}$ = ($\mathcal{V}$, $\mathcal{E}$), where $\mathcal{V}$ and $\mathcal{E}$ are sets of nodes and edges respectively. $\mathbf{E}$ is the similarity matrix with $\mathbf{E}_{i,j}$ as the edge between the i-node $v_i$ and the j-th node $v_j$. Ncut~\cite{shi2000normalized} is proposed to partition the graph into two disjoint sets $\mathcal{A}$ and $\mathcal{B}$. Different to standard graph cut, 
Ncut criterion considers both the total dissimilarity between $\mathcal{A}$ and $\mathcal{B}$ as well as the total similarity within $\mathcal{A}$ and $\mathcal{B}$. Precisely, we seek to minimize the Ncut energy~\cite{shi2000normalized}:

\begin{align}
    \label{eqn:ncut_energy}
	\frac{ \connect(\mathcal{A}, \mathcal{B})}{\connect(\mathcal{A}, \node)} + \frac{\connect (\mathcal{A}, \mathcal{B})}{\connect(\mathcal{B}, \node)},
\end{align}
where $\connect$ measures the degree of similarity between two sets. , $\connect(\mathcal{A}, \mathcal{B}) = \sum_{\feat_i \in \mathcal{A}, \feat_j \in \mathcal{B}} \mathbf{E}_{i,j}$ and $\connect(\mathcal{A}, \node)$ is the total connection from nodes in $\mathcal{A}$ to all the nodes in the graph.

As shown by ~\cite{shi2000normalized}, the equivalent form of optimization problem in Eqn~\ref{eqn:ncut_energy} can be expressed as: 
\begin{align}
	\label{eqn:ncut_final}
	\min_{\indicator} E(\indicator) = \min_{\indicatory} \frac{\indicatory^T (\Degree - \mathbf{E}) \indicatory}{\indicatory^T \Degree \indicatory}\lmm{,}
\end{align}
with the condition of $\indicatory \in \{1, -b\}^N$, where $b$ satisfies $\indicatory^T \Degree \mathbf{1} = 0$, where $\Degree$ is a diagonal matrix with $\degree_i = \sum_{j} \mathbf{E}_{i,j}$ on its diagonal. 

\begin{figure*}[!t]
\begin{tabular}{c@{\hskip 1.3pt}c@{\hskip 1.3pt}c@{\hskip 1.3pt}c@{\hskip 1.3pt}c@{\hskip 1.3pt}c}

		\includegraphics[width=0.16\textwidth, height=0.12\textwidth]{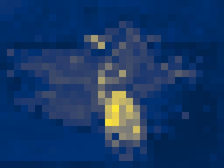} &
		\includegraphics[width=0.16\textwidth, height=0.12\textwidth]{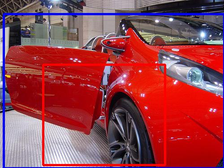} &
		\includegraphics[width=0.16\textwidth, height=0.12\textwidth]{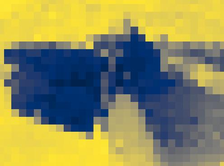} &
        \includegraphics[width=0.16\textwidth, height=0.12\textwidth]{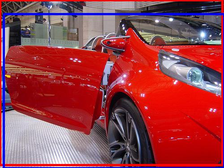} &
		\includegraphics[width=0.16\textwidth, height=0.12\textwidth]{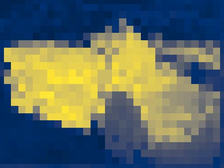} &
		\includegraphics[width=0.16\textwidth, height=0.12\textwidth]{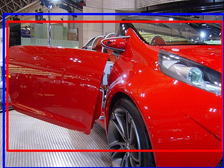} \\ 
		
		\includegraphics[width=0.16\textwidth, height=0.12\textwidth]{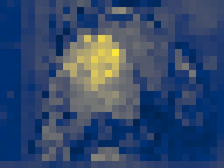} & 
		\includegraphics[width=0.16\textwidth, height=0.12\textwidth]{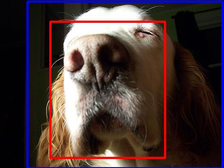} &
		\includegraphics[width=0.16\textwidth, height=0.12\textwidth]{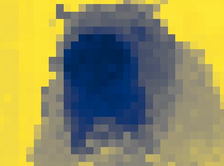} &
		\includegraphics[width=0.16\textwidth, height=0.12\textwidth]{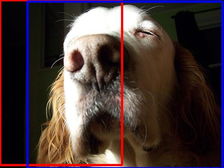} &
		\includegraphics[width=0.16\textwidth, height=0.12\textwidth]{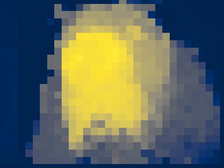} &
		\includegraphics[width=0.16\textwidth, height=0.12\textwidth]{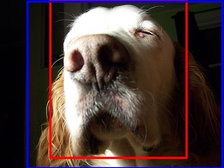} \\
		
		\includegraphics[width=0.16\textwidth, height=0.12\textwidth]{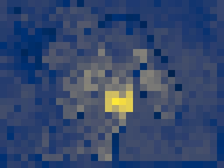} &
		\includegraphics[width=0.16\textwidth, height=0.12\textwidth]{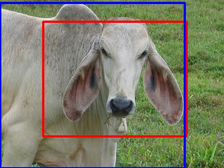} &
		\includegraphics[width=0.16\textwidth, height=0.12\textwidth]{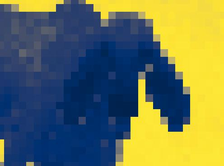} &
		\includegraphics[width=0.16\textwidth, height=0.12\textwidth]{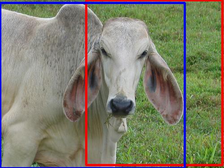} &
		\includegraphics[width=0.16\textwidth, height=0.12\textwidth]{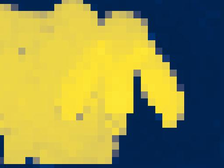} &
		\includegraphics[width=0.16\textwidth, height=0.12\textwidth]{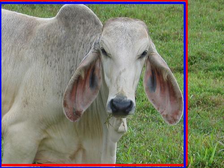} \\
		
        \includegraphics[width=0.16\textwidth, height=0.12\textwidth]{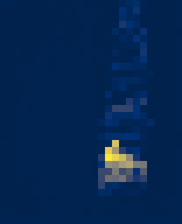} &
		\includegraphics[width=0.16\textwidth, height=0.12\textwidth]{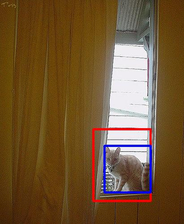} &
        \includegraphics[width=0.16\textwidth, height=0.12\textwidth]{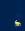} &
		\includegraphics[width=0.16\textwidth, height=0.12\textwidth]{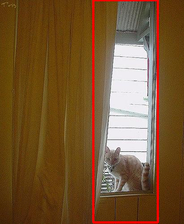} &
		\includegraphics[width=0.16\textwidth, height=0.12\textwidth]{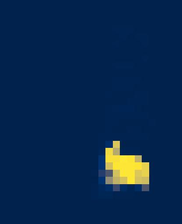} &
		\includegraphics[width=0.16\textwidth, height=0.12\textwidth]{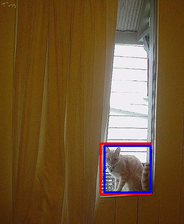} \\
		\makecell{(a) LOST Inverse \\ Degree Attention}  & \makecell{(b) LOST \\ Detection} & \makecell{(c) DSS Eigen\\ Attention} & \makecell{(d) DSS \\ Detection} & \makecell{(e) Our Eigen \\Attention} & \makecell{(f) Our \\Detection} \\
\end{tabular}
\caption{\textbf{Visual results of unsupervised single object discovery on VOC12.} In (a), we show the map of LOST~\cite{simeoni2021localizing} inverse degrees, which is used for detection (b). In (c), DSS~\cite{melaskyriazi2022deep} eigen attention map is shown for its detection in (d), note that DSS is hard to detect large objects leading to an inverse foreground and background in eigen attention map. For our approach, we illustrate the eigenvector in (e) and our detection in (f). \textcolor{blue}{Blue} and \textcolor{red}{Red} bounding boxes indicate the ground-truth and the predicted bounding boxes respectively. }
\label{fig:visual_res}
\end{figure*}

\paragraph*{Ncut solution with the relaxed constraint}
Taking $\indicatorz = \Degree ^{\frac{1}{2}}\indicatory$, Eqn~\ref{eqn:ncut_final} can be rewritten as: 
\begin{align}
	\label{eqn:ncut_z}
	\min_{\indicatorz} \frac{\indicatorz^T \Degree ^{-\frac{1}{2}} (\Degree - \mathbf{E}) \Degree ^{-\frac{1}{2}} \indicatorz}{\indicatorz^T \indicatorz}.
\end{align}
Indicating in~\cite{shi2000normalized}, the formulation in Eqn~\ref{eqn:ncut_z} is equivalent to the Rayleigh quotient~\cite{van1996matrix}, which is equivalent to solve $\Degree ^{-\frac{1}{2}} (\Degree - \mathbf{E}) \Degree ^{-\frac{1}{2}} \indicatorz = \lambda \indicatorz$, where $\Degree - \mathbf{E}$ is the Laplacian matrix and known to be positive semidefinite~\cite{pothen1990partitioning}. Therefore $\indicatorz_0 =  \Degree ^{\frac{1}{2}} \mathbf{1}$ is an eigenvector associated to the smallest eigenvalue $ \lambda = 0$. According to the Rayleigh quotient~\cite{van1996matrix}, the second smallest  eigenvector $\indicatorz_1$ is perpendicular to the smallest one ($\indicatorz_0$) and can be used to minimize the energy in Eqn~\ref{eqn:ncut_z},

\begin{align}
\indicatorz_1 = \argmin_{ \indicatorz^T \indicatorz_0} \frac{\indicatorz^T \Degree ^{-\frac{1}{2}} (\Degree - \mathbf{E}) \Degree ^{-\frac{1}{2}} \indicatorz}{\indicatorz^T \indicatorz}. \nonumber 
\end{align}

Taking $\indicatorz = \Degree ^{\frac{1}{2}}\indicatory$,

\begin{align}
	\indicatory_1 = \argmin_{ \indicatory^T \Degree \mathbf{1} = 0} \frac{\indicatory^T (\Degree - \mathbf{E})  \indicatory}{\indicatory^T \Degree \indicatory}. \nonumber 
\end{align}
Thus, the second smallest eigenvector of the generalized
eigensystem $(\Degree - \mathbf{E})  \indicatory = \lambda  \Degree \indicatory$ provides a solution to the Ncut~\cite{shi2000normalized} problem.

\subsection{The TokenCut Algorithm}
\label{sec:tokencut}

The \name~algorithm consists of three steps: (a) Graph Construction, (b) Graph Cut, (c) Edge Refinement. An overview of the algorithm is shown in Fig.~\ref{fig:method}. 

\subsubsection{Graph construction}
\paragraph*{Image Graph}
\label{sec:image_graph}
As described in Section~\ref{sec:ncut},  \name~operates on a fully connected undirected graph $\graph$~=~($\node$, $\edge$), where $\feat_i$ represents the feature vectors of the node $v_i$. Each patch is linked to other patches by labeled edges,  $\edge$. Edge labels represent a similarity score $\simi$.

\begin{align}
	\label{eqn:edge}
	\edge_{i,j} =
	\begin{cases} 
		1,  & \mbox{if }\simi(\feat_i, \feat_j) \ge \minsimi \\
		\epsilon, & \mbox{else}
	\end{cases},
\end{align}
where $\minsimi$ is a hyper-parameter and $\simi(\feat_i, \feat_j) = \frac{\feat_i \feat_j}{\lVert \feat_i \rVert_2 \lVert \feat_j \rVert_2}$ is the cosine similarity between features. $\epsilon$ is a small value 
$10^{-5}$ to assure a fully connected graph.
Note that the spatial location information has been implicitly included in the features, which is achieved by positional encoding in the transformer. 

\paragraph*{Video Graph} As with images, videos are presented as a fully connected graph where the nodes $\node$ are  visual patches and the edges $\edge$ are labeled with the similarity between patches. However, for videos, similarity includes a score based on both RGB appearance and a RGB representation of optical flow computed between consecutive frames~\cite{baker2011database}. 
The algorithm extracts a sequence of feature vectors using a vision transformer as described in Section~\ref{sec:visiontransformers}. 
Let $\feat^I_i$  and $\feat^F_i$ denote  the feature of i-th image patch and flow patch respectively. Edges are labeled with  the average 
over the similarities between image feature and flow features, expressed as:
\begin{align}
	\label{eqn:edge2}
	\edge_{i,j} =
	\begin{cases} 
		1,  & \mbox{if } \frac{\simi(\feat^I_i, \feat^I_j)+\simi(\feat^F_i, \feat^F_j)}{2}\ge \minsimi \\
		\epsilon, & \mbox{else}
	\end{cases}.
\end{align}

Image feature provide segmentation using appearance while flow features focus on segmentation with motion. We provide a full analysis on the definition of edges in Section~\ref{sec:abl}.

\subsubsection{Graph Cut}
 
The Ncut algorithm is used to partition the fully connected graph. Ncut computes the  second smallest eigenvector of the generalized eigensystem, as described in Section~\ref{sec:ncut} to highlight salient objects. 
We refer to this eigenvector as a measure for ``eigen-attention'', 
and provide visualizations of the attention map provided by this vector in Section~\ref{sec:exp}. 
\name~uses eigen-attention  to bi-partition the graph,  determines which partition belongs to the foreground and then determines the nodes that belong to the each object region.

\paragraph*{Bi-partition the graph}
To partition the nodes into two disjoint sets, \name~uses the average value of the second smallest eigenvector to cut the graph $\overline{\indicatory_1} =  \frac{1}{\nbnode} \sum_{i} \indicatory^i_1$. Formally, $\mathcal{A} = \{\feat_i | \indicatory^i_1 \le \overline{\indicatory_1} \}$ and $\mathcal{B} = \{\feat_i | \indicatory^i_1 > \overline{\indicatory_1} \}$. Note that, we also explored the use of classical clustering algorithms, such as K-means and EM, to cluster the second smallest eigenvector into 2 partitions. The comparison is available in Section~\ref{sec:abl}. Our experimens show that the average value generally provides better results.

\begin{table*}[!t]
\begin{center}
    \caption{\textbf{Comparisons for unsupervised single object discovery}. We compare \name~to state-of-the-art object discovery methods on VOC07~\cite{pascal-voc-2007}, VOC12 ~\cite{pascal-voc-2012} and COCO20K~\cite{lin2014microsoft,vo2020toward} datasets. Model performances are evaluated with CorLoc metric. ``Inter-image Simi.'' means the model leverages information from the entire dataset and explores inter-image similarities to localize objects.}
    \label{tab: object_dis}
\resizebox{1.8\columnwidth}{!}{

\begin{tabular}{lcllll}
    \toprule Method & Inter-image Simi.& DINO~\cite{caron2021emerging} Feat. & VOC07~\cite{pascal-voc-2007} & VOC12 ~\cite{pascal-voc-2012}& COCO20K~\cite{lin2014microsoft,vo2020toward}  \\
    \midrule
    Selective Search~\cite{uijlings2013selective, simeoni2021localizing}&  &  - & 18.8 & 20.9 & 16.0 \\
    EdgeBoxes~\cite{zitnick2014edge, simeoni2021localizing} &  & -& 31.1 & 31.6 & 28.8 \\
    Kim et al.~\cite{kim2009unsupervised, simeoni2021localizing}& \cmark  & -& 43.9 & 46.4& 35.1 \\
    Zhange et al.~\cite{zhang2020object, simeoni2021localizing}& \cmark & -& 46.2 & 50.5 & 34.8 \\
    DDT+~\cite{wei2019unsupervised, simeoni2021localizing}& \cmark & -& 50.2 & 53.1 & 38.2 \\
    rOSD~\cite{vo2020toward, simeoni2021localizing} &\cmark & -&  54.5 & 55.3 & 48.5 \\
    LOD~\cite{vo2021large, simeoni2021localizing}&\cmark &  -& 53.6 & 55.1 & 48.5 \\
    DINO-seg~\cite{caron2021emerging,simeoni2021localizing}&  & ViT-S/16~\cite{dosovitskiy2020image} &  45.8 & 46.2 & 42.1 \\
    LOST~\cite{simeoni2021localizing}&   &  ViT-S/16~\cite{dosovitskiy2020image} & 61.9 & 64.0 & 50.7 \\
    DSS~\cite{melaskyriazi2022deep}&   &  ViT-S/16~\cite{dosovitskiy2020image} & 62.7 & 66.4 & 56.2 \\
    
    \bf TokenCut &  &  ViT-S/16~\cite{dosovitskiy2020image} & \bf 68.8 (\textcolor{cssgreen}{$\uparrow$ \bf 6.1}) &  \bf 72.1 (\textcolor{cssgreen}{$\uparrow$ \bf 5.7}) &  \bf 58.8 (\textcolor{cssgreen}{$\uparrow$ \bf 2.6})\\ 
    \midrule
    LOD + CAD$^{\star}$~\cite{simeoni2021localizing} & \cmark & -& 56.3 & 61.6 & 52.7 \\
    rOSD + CAD$^{\star}$~\cite{simeoni2021localizing} & \cmark & -& 58.3 & 62.3 & 53.0 \\
    LOST + CAD$^{\star}$~\cite{simeoni2021localizing} &  & ViT-S/16~\cite{dosovitskiy2020image}& 65.7 & 70.4 & 57.5 \\
    \bf TokenCut + CAD$^{\star}$~\cite{simeoni2021localizing} &  & ViT-S/16~\cite{dosovitskiy2020image}& \bf 71.4 (\textcolor{cssgreen}{$\uparrow$ \bf 5.7}) & \bf 75.3 (\textcolor{cssgreen}{$\uparrow$ \bf 4.9})&  \bf 62.6 (\textcolor{cssgreen}{$\uparrow$ \bf 5.1}) \\
\bottomrule
\end{tabular}}
\end{center}
\begin{center}
\vspace{1pt}
  \footnotesize{$^{\star}$ +CAD indicates to train a second stage class-agnostic detector with ``pseudo-boxes'' labels.} 
\end{center}

\end{table*}

\paragraph*{Foreground Determination}
Given the two disjoint sets of nodes, \name~selects the partition with the maximum absolute value $\feat_{max}$ as the foreground. 
Intuitively, the foreground object should be salient and thus less connected to the entire graph. In other words, $\degree_i < \degree_j$ if $\feat_i$ belongs to the foreground while $\feat_j$ is the background token. Therefore, the eigenvector of the foreground object should have a larger absolute value than the  background region. 

\paragraph*{Select the object}
In images, we are interested in segmenting a single object. However,  the foreground can contain more than one salient object region. \name~selects the connected component in the foreground containing the maximum absolute value $\feat_{max}$ as the detected object. In videos, as the goal is to segment objects based on both motion and appearance, \name~takes the entire foreground region as the final output.

\subsubsection{Edge Refinement}

The graph cut algorithm provides coarse masks of object regions due to the large size of transformer patches. The boundaries of such masks can be easily refined using standard edge refinement technique. We have experimented with off-the-shelf edge-aware post-processing techniques such as Bilateral Solver~\cite{barron2016fast} (BS), Conditional Random Field~\cite{krahenbuhl2011efficient} (CRF) on top of the obtained coarse mask to generate more precise boundaries for the mask. We have found that CRF usually provides the best results.

\hfill \break
\section{Experiments}
\label{sec:exp}
We evaluated the suitability of \name~for three tasks: unsupervised single object discovery, unsupervised saliency detection and unsupervised video segmentation. We present implementation details in Section ~\ref{sec:implementation_details}. The results of unsupervised single object discovery are shown in Section~\ref{sec:unsupervised_object_discovery}.  The results for unsupervised saliency detection are presented in Section~\ref{sec:unsupervised_saliency_detection}, and results for unsupervised video segmentation in Section~\ref{sec:UVS}. We provide ablation studies in Section~\ref{sec:abl}.

\subsection{Implementation details}
\label{sec:implementation_details}

\paragraph*{Model details}
For our experiments, we use the ViT-S/16 model~\cite{dosovitskiy2020image} trained with self-distillation loss (DINO)~\cite{caron2021emerging} to extract features of patches. Following~\cite{simeoni2021localizing}, we employ the key features of the last layer as the input features $\feat$. Ablations on different features and ViT backbones are  provided in Tab.~\ref{tab:backbone}. We set $\tau = 0.2$ for all image datasets and $\tau = 0.3$ for video datasets. The selection of $\tau$ is discussed in Section~\ref{sec:abl}.

\paragraph*{Algorithmic Cost}
In terms of running time, our implementation takes approximately 0.32 seconds to detect a bounding box for a salient object region in a single image with resolution 480 $\times$ 480 using a single GPU QUADRO RTX 8000. Obtainaing a coarse mask from 20 frames of video with 320 x 576 resolution, requires an average of 30 seconds with standard deviation of around 4.5 seconds. 
Edge refinement the takes an additional 16.4 seconds on average with a standard deviation 1.4 seconds. As with single-frame graphs, the same video takes 0.93 seconds in average to obtain the coarse mask with standard deviation of 0.17 for all frames. The post processing step cost 16.1 seconds with standard deviation of 1.4.   For n tokens, the algorithmic complexity for building such a graph  is $O(n^2)$. Thus the average processing time grows with the square of the number of frames in the video.

\paragraph*{Optical flow details}
To generate optical flow, we use two different approaches: RAFT~\cite{teed2020raft} and ARFlow~\cite{liu2020learning}. The first one is supervised and the second one is self-supervised. We extract the optical ﬂow at the original resolution of the image pairs, with the frame gaps n = 1 for DAVIS~\cite{perazzi2016benchmark} and SegTV2~\cite{li2013video} dataset. For FBMS~\cite{ochs2013segmentation} we use n = 3 to compensate for the much slower rate of motion. This improves the optical flow quality as small pixel-level motions are hard to detect using off-the-shelf methods. Optical flow features are encoded as RGB values, using standard techniques for visualization of optical flow~\cite{baker2011database}. This allows us to directly use the pre-trained self-supervised transformers with optical flow encoded as RGB. 
Because of limits on available  computational resources, we construct the video graph with a maximum of 90 frames on the DAVIS dataset. For videos longer than 90 frames, it is possible to aggregate results using non-overlapping subgraphs with maximum video frames of 90. %

\subsection{Unsupervised Single Object Discovery}
\label{sec:unsupervised_object_discovery}

\begin{table*}[!ht]
\centering
\caption{\textbf{Comparisons for unsupervised saliency detection} We compare \name~to state-of-the-art unsupervised saliency detection methods on ECSSD~\cite{shi2015hierarchical}, DUTS~\cite{wang2017learning} and DUT-OMRON~\cite{yang2013saliency}. \name~achieves better results compared with other competitive approaches.}

\resizebox{1\textwidth}{!}{
\begin{tabular}{l|lll|lll|lll}
\toprule

\multirow{2}{*}{Method} & \multicolumn{3}{c|}{ECSSD~\cite{shi2015hierarchical}} & \multicolumn{3}{c|}{DUTS~\cite{wang2017learning}}  & \multicolumn{3}{c}{DUT-OMRON~\cite{yang2013saliency}} \\
                        & \multicolumn{1}{l}{$maxF_{\beta}$(\%)} & \multicolumn{1}{l}{IoU(\%)} & \multicolumn{1}{l|}{Acc.(\%)} & \multicolumn{1}{l}{$maxF_{\beta}$(\%)} & \multicolumn{1}{l}{IoU(\%)} & \multicolumn{1}{l|}{Acc.(\%)} & \multicolumn{1}{l}{$maxF_{\beta}$(\%)} & \multicolumn{1}{l}{IoU(\%)} & \multicolumn{1}{l}{Acc.(\%)} \\
\midrule
HS~\cite{yan2013hierarchical}  & 67.3  & 50.8  & 84.7  & 50.4      & 36.9   & 82.6  & 56.1  & 43.3   & 84.3 \\
wCtr~\cite{zhu2014saliency}    & 68.4  & 51.7  & 86.2  & 52.2      & 39.2   & 83.5  & 54.1  & 41.6   & 83.8 \\
WSC~\cite{li2015weighted}      & 68.3  & 49.8  & 85.2  & 52.8      & 38.4   & 86.2  & 52.3  & 38.7   & 86.5 \\
DeepUSPS~\cite{nguyen2019deepusps} & 58.4  & 44.0  & 79.5  & 42.5      & 30.5   & 77.3  & 41.4  & 30.5   & 77.9 \\
BigBiGAN~\cite{voynov2021object}  & 78.2  & 67.2  & 89.9  & 60.8      & 49.8   & 87.8  & 54.9  & 45.3   & 85.6 \\
E-BigBiGAN~\cite{voynov2021object}  & 79.7  & 68.4  & 90.6  & 62.4      & 51.1   & 88.2  & 56.3  & 46.4   & 86.0 \\
LOST~\cite{simeoni2021localizing,shen2021learning}   & 75.8  & 65.4  & 89.5  & 61.1      & 51.8   & 87.1  & 47.3  & 41.0   & 79.7 \\
LOST~\cite{simeoni2021localizing,shen2021learning}+BS~\cite{barron2016fast}   & 83.7  & 72.3  & 91.6  & 69.7      & 57.2   & 88.7  & 57.8  & 48.9   & 81.8 \\           
DSS~\cite{melaskyriazi2022deep} & - & 73.3  & - & - & 51.4 & - & - & 56.7 & - \\
\midrule
\bf TokenCut & 80.3	& 71.2	& 91.8	& 67.2 &	57.6 & 	90.3 &	60.0 &	53.3 &	88.0 \\
\bf TokenCut + BS~\cite{barron2016fast} & \bf 87.4 (\textcolor{cssgreen}{$\uparrow$ \bf 3.7}) &  77.2 & 93.4 &  75.5 &	 62.4 &  91.4 & \bf 69.7 (\textcolor{cssgreen}{$\uparrow$ \bf 11.9}) &  61.8 &  89.7 \\
\bf TokenCut + CRF~\cite{krahenbuhl2011efficient} & \bf 87.4 (\textcolor{cssgreen}{$\uparrow$ \bf 3.7})	& \bf 77.7 (\textcolor{cssgreen}{$\uparrow$ \bf 4.4})	& \bf93.6 (\textcolor{cssgreen}{$\uparrow$ \bf 2.0})	& \bf 75.7 (\textcolor{cssgreen}{$\uparrow$ \bf 6.0}) &	\bf 62.8 (\textcolor{cssgreen}{$\uparrow$ \bf 5.6})& \bf 91.5 (\textcolor{cssgreen}{$\uparrow$ \bf 2.8}) & 69.2 &	\bf 61.9 (\textcolor{cssgreen}{$\uparrow$ \bf 5.2}) & \bf 89.8 (\textcolor{cssgreen}{$\uparrow$ \bf 8.0}) \\

\bottomrule
\end{tabular}
}
\label{tab:salient_detection}
\end{table*}

\paragraph*{Datasets}
\name~has been evaluated on three commonly used benchmarks for unsupervised single object discovery: VOC07~\cite{pascal-voc-2007}, VOC12~\cite{pascal-voc-2012} and COCO20K ~\cite{lin2014microsoft,vo2020toward}. 
VOC07 and VOC12 contain 5011 and 11540 images respectively which belong to 20 categories. COCO20K consists of 19817 randomly chosen images from the COCO2014 dataset~\cite{lin2014microsoft}. VOC07 and VOC12 are commonly used to evaluate unsupervised object discovery~\cite{vo2020toward,vo2021large,vo2019unsupervised, wei2019unsupervised,cho2015unsupervised}. COCO20K is a popular benchmark for a large scale evaluation~\cite{vo2020toward}.

\paragraph*{Evaluation metric}
In line with previous research~\cite{deselaers2010localizing, vo2020toward,vo2021large,vo2019unsupervised, wei2019unsupervised,cho2015unsupervised,siva2013looking},
we report performance using the \emph{CorLoc} metric for precise localization. We use a single predicted bounding box for each image. For target images, CorLoc is 1.0 if the intersection over union (IoU) score between the predicted bounding box and the ground truth bounding boxes is superior to 0.5.

\begin{figure*}[!t]
\resizebox{\linewidth}{!}{ \begin{tabular}{c@{\hskip 1.3pt}c@{\hskip 1.3pt}c@{\hskip 1.3pt}c@{\hskip 1.3pt}c@{\hskip 1.3pt}c@{\hskip 1.3pt}c}
        
		\rotatebox{90}{\makecell{\small ~~~~~~~(a)~GT}} &
		\includegraphics[width=0.16\textwidth, height=0.12\textwidth]{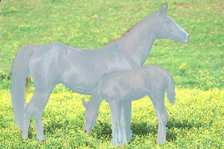} &
		\includegraphics[width=0.16\textwidth, height=0.12\textwidth]{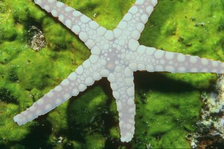} &
		\includegraphics[width=0.16\textwidth, height=0.12\textwidth]{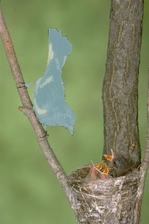} &
    	\includegraphics[width=0.16\textwidth, height=0.12\textwidth]{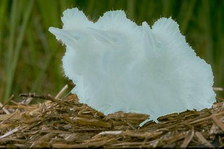} &
    	\includegraphics[width=0.16\textwidth, height=0.12\textwidth]{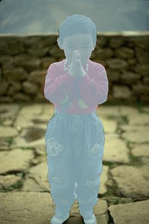} &
		\includegraphics[width=0.16\textwidth, height=0.12\textwidth]{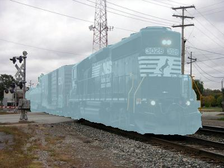} \\
		
		\rotatebox{90}{\makecell{\small ~~~(b)~\name}} &
		\includegraphics[width=0.16\textwidth, height=0.12\textwidth]{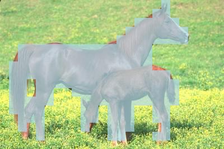} &
		\includegraphics[width=0.16\textwidth, height=0.12\textwidth]{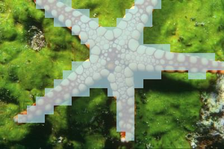} &
		\includegraphics[width=0.16\textwidth, height=0.12\textwidth]{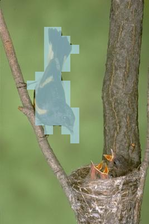} &
		\includegraphics[width=0.16\textwidth, height=0.12\textwidth]{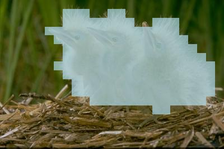} &
		\includegraphics[width=0.16\textwidth, height=0.12\textwidth]{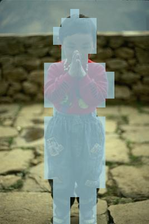} &
		\includegraphics[width=0.16\textwidth, height=0.12\textwidth]{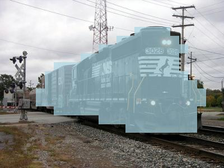} \\
		
		\rotatebox{90}{\makecell{\small(c)~\name \\ \small+ CRF}} &
		\includegraphics[width=0.16\textwidth, height=0.12\textwidth]{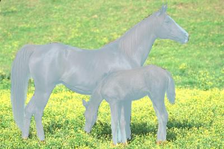} &
		\includegraphics[width=0.16\textwidth, height=0.12\textwidth]{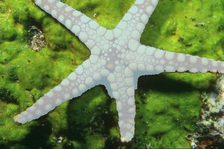} &
		\includegraphics[width=0.16\textwidth, height=0.12\textwidth]{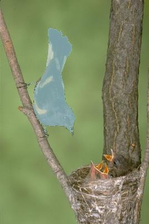} &
		\includegraphics[width=0.16\textwidth, height=0.12\textwidth]{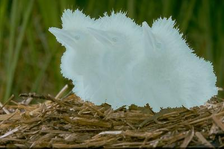} &
		\includegraphics[width=0.16\textwidth, height=0.12\textwidth]{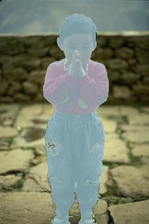} &
		\includegraphics[width=0.16\textwidth, height=0.12\textwidth]{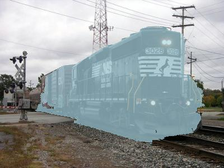} \\

\end{tabular}}

\caption{\textbf{Visual results of unsupervised segments on ECSSD~\cite{shi2015hierarchical}}. In (a), we show the ground truth. (b) is {\name} coarse mask segmentation result. The performance of {\name} + Bilateral Solver (BS) is presented in (c).} 
\label{fig: saliency}
\end{figure*}

\paragraph*{Quantitative Results}
We evaluate the CorLoc scores in comparison with previous state-of-the-art single object discovery methods~\cite{uijlings2013selective, zitnick2014edge, simeoni2021localizing, kim2009unsupervised, zhang2020object, wei2019unsupervised, vo2020toward, vo2021large} on VOC07, VOC12, and COCO20K datasets. These methods can be roughly divided into two groups \CUT{based on} according to whether the model uses information from the entire dataset or explores inter-image similarities. Because of the quadratic complexity of region comparison among images, models with inter-image similarities are generally difficult to scale to larger datasets. The selective search~\cite{uijlings2013selective}, edge boxes~\cite{zitnick2014edge}, LOST~\cite{simeoni2021localizing} and {\name} do not require inter-image similarities and are thus much more efficient. As shown in the Tab.~\ref{tab: object_dis},  {\name} consistently outperforms all the previous methods on all the datasets by a large margin. 
Particularly, {\name} ouperforms DSS~\cite{melaskyriazi2022deep} by 6.1\%, 5.7\% and 2.6\% for VOC07, VOC12 and COCO20K respectively using the same ViT-S/16 features.

We also list a set of results that includes using a second stage unsupervised training strategy to boost the performance. This is referred to as Class-Agnostic Detection (CAD) and proposed in LOST~\cite{simeoni2021localizing}. For this, we first compute K-means on all the boxes produced by the first stage single object discovery model to obtain pseudo labels of the bounding boxes. Then a classical Faster RCNN~\cite{ren2015faster} is trained on the pseudo labels. As shown in Tab.~\ref{tab: object_dis}, {\name} with CAD outperforms the state-of-the-art by 5.7\%, 4.9\% and 5.1\% on VOC07, VOC12 and COCO20k respectively.

\begin{figure}[!t]

\centering
\begin{tabular}{c@{\hskip 3pt}c@{\hskip 3pt}c}
        \includegraphics[width=0.15\textwidth, height=0.13\textwidth]{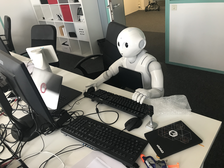} &
        \includegraphics[width=0.15\textwidth, height=0.13\textwidth]{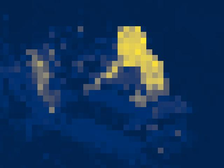}&
        \includegraphics[width=0.15\textwidth, height=0.13\textwidth]{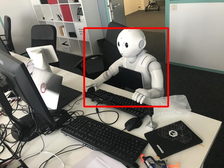} \\

        \includegraphics[width=0.15\textwidth, height=0.13\textwidth]{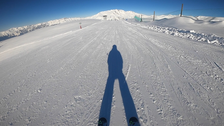} &
        \includegraphics[width=0.15\textwidth, height=0.13\textwidth]{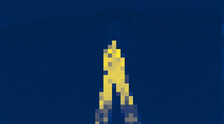}&
        \includegraphics[width=0.15\textwidth, height=0.13\textwidth]{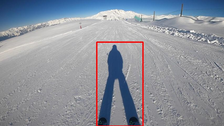} \\	
		
        \includegraphics[width=0.15\textwidth, height=0.13\textwidth]{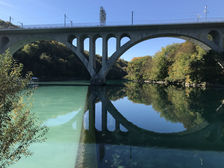} &
        \includegraphics[width=0.15\textwidth, height=0.13\textwidth]{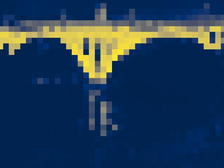}&
        \includegraphics[width=0.15\textwidth, height=0.13\textwidth]{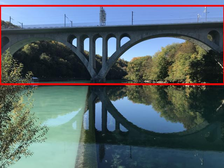} \\
        
        \includegraphics[width=0.15\textwidth, height=0.13\textwidth]{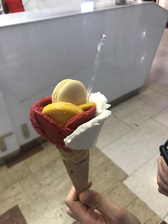} &
        \includegraphics[width=0.15\textwidth, height=0.13\textwidth]{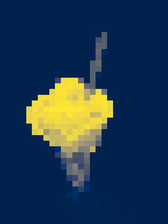}&
        \includegraphics[width=0.15\textwidth, height=0.13\textwidth]{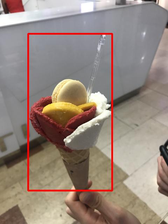} \\
        
        (a) Input & (b) Eigen Attention & (c) Detection \\ 
\end{tabular}

\caption{\textbf{Visualization of images from the Internet.} We show the input images, our eigen attention, and final detection in (a), (b), and (c) respectively.}
\label{fig:extra_samples}

\end{figure}

\paragraph*{Qualitative Results}
 In Fig.~\ref{fig:visual_res}, we provide visualization for LOST~\cite{simeoni2021localizing}, DSS~\cite{melaskyriazi2022deep} and  {\name}\footnote{More visual results can be found in the \href{https://www.m-psi.fr/Papers/TokenCut2022/}{project webpage}.}. For each method, we visualize the heatmap that is used to perform object detection. For LOST, the detection is mainly based on the map of inverse degree ($\frac{1}{\degree_i}$). For DSS, the heatmap is the attention map associated to the second eigenvector. For {\name}, we display the second smallest eigenvector.  The visual results demonstrate that  {\name} can extract a high quality segmentation for the salient object. Compared with LOST and DSS, {\name} is able to extract a more complete segmentation as can be seen in the first and the second samples in Fig.~\ref{fig:visual_res}. In other cases, when LOST and DSS are unable to detect a large object, {\name} can detect the object properly. Examples for this can be seen in the third and fourth samples in Fig.~\ref{fig:visual_res}. 

\paragraph*{Internet Images} We further tested {\name} on Internet images\footnote{We provide an \href{https://huggingface.co/spaces/yangtaowang/TokenCut}{online demo} allowing to test Internet images.}. The results are in Fig~\ref{fig:extra_samples}.  It can be seen that even though the input images have noisy backgrounds, {\name} can  provide a precise attention map to cover the object and lead to an accurate prediction of the bounding box, demonstrates robustness of the method.

\begin{figure*}[!t]
\begin{minipage}{\linewidth}
\begin{center}
\resizebox{\linewidth}{!}{ 
\begin{tabular}{c@{\hskip 3pt}c@{\hskip 3pt}c@{\hskip 3pt}c@{\hskip 3pt}c@{\hskip 3pt}c@{\hskip 3pt}c}

\rotatebox{90}{\makecell{~~~~~~(a)~GT}}& 
\includegraphics[width=0.16\textwidth, height=0.12\textwidth]{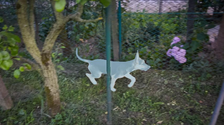} &
\includegraphics[width=0.16\textwidth, height=0.12\textwidth]{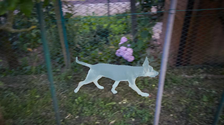} & 
\includegraphics[width=0.16\textwidth, height=0.12\textwidth]{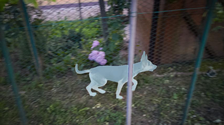} & 
\includegraphics[width=0.16\textwidth, height=0.12\textwidth]{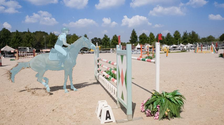} &
\includegraphics[width=0.16\textwidth, height=0.12\textwidth]{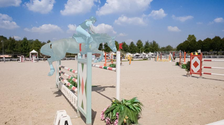} &
\includegraphics[width=0.16\textwidth, height=0.12\textwidth]{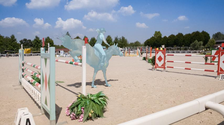} \\

\rotatebox{90}{\makecell{~~~~~(b)~Ours}} &  
\includegraphics[width=0.16\textwidth, height=0.12\textwidth]{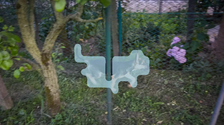} &
\includegraphics[width=0.16\textwidth, height=0.12\textwidth]{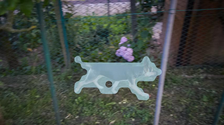} & 
\includegraphics[width=0.16\textwidth, height=0.12\textwidth]{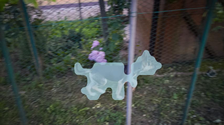} & 
\includegraphics[width=0.16\textwidth, height=0.12\textwidth]{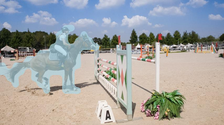} &
\includegraphics[width=0.16\textwidth, height=0.12\textwidth]{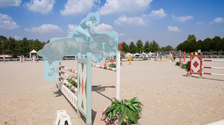} &
\includegraphics[width=0.16\textwidth, height=0.12\textwidth]{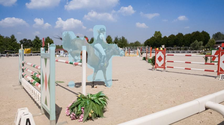} \\

\rotatebox{90}{\makecell{(c)~Ours + CRF}} &
\includegraphics[width=0.16\textwidth, height=0.12\textwidth]{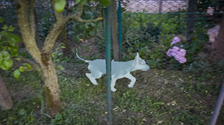} &
\includegraphics[width=0.16\textwidth, height=0.12\textwidth]{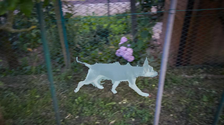} & 
\includegraphics[width=0.16\textwidth, height=0.12\textwidth]{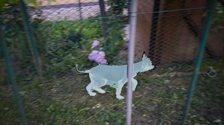} & 
\includegraphics[width=0.16\textwidth, height=0.12\textwidth]{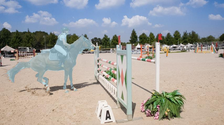} &
\includegraphics[width=0.16\textwidth, height=0.12\textwidth]{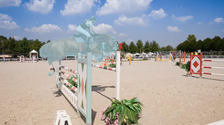} &
\includegraphics[width=0.16\textwidth, height=0.12\textwidth]{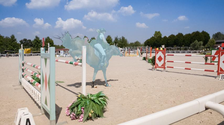} \\
\end{tabular}}
\caption{\textbf{Visual results of unsupervised video segmentation on DAVIS~\cite{perazzi2016benchmark}.} In (a), we show the ground truth segmentation. For \name, we illustrate its coarse mask in (b) and refinement results with CRF in (c).}
\label{fig: davis}
\end{center}
\end{minipage}
\end{figure*}

\subsection{Unsupervised Saliency detection}
\label{sec:unsupervised_saliency_detection}

\paragraph*{Datasets}
We validated the performance of \name for unsupervised Saliency detection using three datasets : Extended Complex Scene Saliency Dataset(ECSSD)~\cite{shi2015hierarchical}, DUTS~\cite{wang2017learning} and DUT-OMRON~\cite{yang2013saliency}. ECSSD contains 1 000 real-world images of complex scenes for testing. 
DUTS contains 10 553 train and 5 019 test images. The training set is collected from the ImageNet detection train/val set. The test set is collected from ImageNet test, and the SUN dataset~\cite{Xiao2010Sun}. Following the previous work~\cite{shen2021learning}, we report the performance on the DUTS-test subset. DUT-OMRON~\cite{yang2013saliency} contains 5 168 images of high quality natural images for testing.

\paragraph*{Evaluation Metrics}
We report three standard metrics: F-measure, IoU and Accuracy. \textit{F-measure} is a standard measure for saliency detection, computed as $F_\beta = \frac{(1+\beta^2)Precision \times Recall}{\beta^2Precision + Recall}$, where the Precision and Recall are defined using a binarized predicted mask and a ground truth mask. The $maxF_{\beta}$ is the maximum value of 255 uniformly distributed binarization thresholds. Following previous work~\cite{shen2021learning,voynov2021object}, we set $\beta=0.3$ for consistency. \textit{IoU}(Intersection over Union) score is computed based on the binary predicted mask and the ground-truth, the threshold is set to 0.5. \textit{Accuracy} measures the proportion of pixels that have been correctly assigned to the object/background. The binarization threshold is set to 0.5 for masks.

\paragraph*{Results}
Qualitative results are shown in Tab.~\ref{tab:salient_detection}. \name~significantly outperforms previous state-of-the-art methods. Adding BS~\cite{barron2016fast} or CRF~\cite{krahenbuhl2011efficient} refines the boundary of an object and further boosts the \name~performance, as can  be seen in the visual results presented in Fig.~\ref{fig: saliency}.

\begin{table}[!t]

\begin{center}
\caption{\textbf{Comparisons for unsupervised video segmentation}. We report \textit{Jaccard index} and compare \name~to state-of-the-art unsupervised video segmentation methods on DAVIS~\cite{perazzi2016benchmark}, FBMS~\cite{ochs2013segmentation} and SegTV2~\cite{li2013video}. \name~achieves  results that are similar to competing approaches. }
\resizebox{\linewidth}{!}{
\begin{tabular}{cccccc}
\toprule
Method & Flow &Training    & DAVIS~\cite{perazzi2016benchmark} & FBMS~\cite{ochs2013segmentation}  & SegTV2~\cite{li2013video}                   \\
\midrule
ARP~\cite{koh2017primary}& CPM~\cite{hu2016efficient}&  & 76.2  & 59.8  & 57.2                     \\
ELM~\cite{lao2018extending}& Classic+NL~\cite{sun2010secrets}&                & 61.8  & 61.6  & -                        \\
MG~\cite{yang2021self}  & RAFT~\cite{teed2020raft}& \cmark      & 68.3  & 53.1  & 58.2                     \\
CIS~\cite{yang2019unsupervised}& PWCNet~\cite{sun2018pwc}& \cmark     & 71.5  & 63.5  & 62                       \\
DyStaB~\cite{yang2021dystab}$^{\star}$ & PWCNet~\cite{sun2018pwc}& \cmark           & \bf 80.0    & \bf 73.2  & \bf 74.2                     \\
DeSprite~\cite{ye2022deformable}$^{\ddagger}$ & RAFT~\cite{teed2020raft}&  \cmark& 79.1  & 71.8  & 72.1                     \\
\midrule
\textbf{TokenCut} &RAFT~\cite{teed2020raft}&  & 64.3  & 60.2 &  59.6 \\
\textbf{TokenCut + BS~\cite{barron2016fast}} &RAFT~\cite{teed2020raft}&  & 75.1  & 61.2 & 56.4 \\
\textbf{TokenCut + CRF~\cite{krahenbuhl2011efficient}}&RAFT~\cite{teed2020raft}&  & 76.7  & 66.6 & 61.6 \\
\textbf{TokenCut} &ARFlow~\cite{liu2020learning}&  & 62.0  & 61.0 & 58.9 \\
\textbf{TokenCut + BS~\cite{barron2016fast}} &ARFlow~\cite{liu2020learning}&  &  73.1 & 64.7& 54.6 \\
\textbf{TokenCut + CRF~\cite{krahenbuhl2011efficient}}&ARFlow~\cite{liu2020learning}&  & 74.4  & 69.0 & 60.8 \\

\bottomrule
\end{tabular}}

\vspace{2pt}

\footnotesize{$\star$:~\cite{yang2021dystab} is trained on DAVIS and evaluated on FBMS and SegTV2; \\ $\ddagger$:~\cite{ye2022deformable} is optimized for each video separately.}
\label{tab:uvs}
\end{center}
\end{table}

\subsection{Unsupervised Video Segmentation}
\label{sec:UVS}
\paragraph*{Datasets}
We further evaluate \name~using three commonly used datasets for unsupervised video segmentation: DAVIS~\cite{perazzi2016benchmark}, FBMS~\cite{ochs2013segmentation} and SegTV2~\cite{li2013video}.
DAVIS contains 50 high-resolution real-word videos, where 30 are for training and 20 are for validation. Pixel-wise annotations are depicted for the principle moving object within the scene for each frame. FBMS consists of 59 multiple moving object videos, providing 30 videos for testing with a total of 720 annotation frames. SegTV2 contains 14 full pixel-level annotated video for multiple objects segmentation. Following~\cite{yang2021self}, we fuse the annotation of all moving objects into a single mask on FBMS and SegTV2 datasets for fair comparison.

\paragraph*{Evaluation metrics} 
We report performance using \textit{Jaccard index}. The Jaccard index measures the intersection of union between an output segmentation M and the corresponding ground-truth mask G, which has been formulated as $\J = \frac{\left|M \cap G\right|}{\left|M \cup G\right|}$.

\paragraph*{Results}
We compare \name~to the state-of-the art unsupervised video segmentation results in Tab.~\ref{tab:uvs}. \name~achieves competitive performances for this task. Note that DyStaB~\cite{yang2021dystab} must be trained on the entire DAVIS training set and uses the pretrained model for evaluation with the FBMS and SegTV2 datasets. DeSprite~\cite{ye2022deformable} learns an auto-encoder model to optimize on each individual video. In contrast,  \name~does not require training  and generalizes well for all three datasets. Visual results are illustrated in Fig.~\ref{fig: davis}, \name~can precisely segment moving objects even in the case of challenging occlusions. Adding CRF as a post-processing further improves  the boundary for segmented regions\footnote{The segmentation results of entire videos can be found in the ~\href{https://www.m-psi.fr/Papers/TokenCut2022/}{project webpage}.}.

\subsection{Analysis}
\label{sec:abl}

\begin{table}[!t]
	\begin{center}
	\caption{\textbf{Analysis of $\tau$.} We report CorLoc for unsupervised single object discovery on VOC07~\cite{pascal-voc-2007}, VOC12 ~\cite{pascal-voc-2012} and COCO20K~\cite{lin2014microsoft,vo2020toward} datasets, and Jacard index on DAVIS~\cite{perazzi2016benchmark}.}  
 	
	\resizebox{\columnwidth}{!}{
    \begin{tabular}{c | ccc | c}
    \toprule \multirow{2}{*}{$\tau$} & \multicolumn{3}{c|}{CorLoc}& \multicolumn{1}{c}{Jaccard Index}\\
    & VOC07~\cite{pascal-voc-2007} & VOC12 ~\cite{pascal-voc-2012}& COCO20K~\cite{lin2014microsoft,vo2020toward} & DAVIS~\cite{perazzi2016benchmark} \\
    \midrule
     0 &   67.4 & 71.3  & 56.1 & 70.7\\
     0.1 &   68.6  & 72.1  & 58.2 & 74.6\\
     0.2 & \bf 68.8  & \bf 72.1  & \bf 58.8 & 75.8 \\
     0.3 & 67.7  & 72.1  &  58.2 & \bf 76.7\\
    \bottomrule
    \end{tabular}}
	\label{tab:tau}
	\end{center}
\end{table}

\begin{table}[!t]
	\begin{center}
	\caption{\textbf{Analysis of different backbones.} We report CorLoc for unsupervised single object discovery on VOC07~\cite{pascal-voc-2007}, VOC12~\cite{pascal-voc-2012} and COCO20K~\cite{lin2014microsoft,vo2020toward} datasets.}   
	\resizebox{\columnwidth}{!}{
    \begin{tabular}{llllll}
    \toprule Method & Backbone & VOC07~\cite{pascal-voc-2007} & VOC12~\cite{pascal-voc-2012} & COCO20K~\cite{lin2014microsoft,vo2020toward}\\
    \midrule
    LOST~\cite{simeoni2021localizing} &  DINO-S/16~\cite{dosovitskiy2020image,caron2021emerging} & 61.9 & 64.0 & 50.7 \\
    \bf TokenCut &   DeiT-S/16~\cite{dosovitskiy2020image,touvron2021training} &  2.39  & 2.9  & 3.5\\ 
    \bf TokenCut &   MoCoV3-S/16~\cite{dosovitskiy2020image,chen2021empirical} &  66.2  &  66.9 & 54.5\\ 
    \bf TokenCut &   DINO-S/16~\cite{dosovitskiy2020image,caron2021emerging} & \bf 68.8 (\textcolor{cssgreen}{$\uparrow$ \bf 6.9}) &  72.1 (\textcolor{cssgreen}{$\uparrow$ \bf 8.1}) &  58.8 (\textcolor{cssgreen}{$\uparrow$ \bf 8.1})\\ 
    \midrule 
    LOST~\cite{simeoni2021localizing}& DINO-S/8~\cite{dosovitskiy2020image,caron2021emerging} & 55.5 & 57.0 & 49.5\\
    \bf TokenCut  & DINO-S/8~\cite{dosovitskiy2020image,caron2021emerging} &  67.3 (\textcolor{cssgreen}{$\uparrow$ \bf 11.8})    &  71.6 (\textcolor{cssgreen}{$\uparrow$ \bf 14.6}) &  \bf 60.7 (\textcolor{cssgreen}{$\uparrow$ \bf 11.2})\\ 
    \midrule
    LOST~\cite{simeoni2021localizing}& DINO-B/16~\cite{dosovitskiy2020image,caron2021emerging} & 60.1 & 63.3 & 50.0 \\
    \bf TokenCut& MAE-B/16~\cite{dosovitskiy2020image, he2022masked} & 61.5  & 67.4 & 47.7 \\
    \bf TokenCut  & DINO-B/16~\cite{dosovitskiy2020image, caron2021emerging} &  68.8 (\textcolor{cssgreen}{$\uparrow$ \bf 8.7})    &  \bf 72.4 (\textcolor{cssgreen}{$\uparrow$ \bf 9.1}) & 59.0 (\textcolor{cssgreen}{$\uparrow$ \bf 9.0})\\
    \bottomrule
    \end{tabular}}
   
	\label{tab:backbone}
		
	\end{center}

\end{table} 

\paragraph*{Impact of $\tau$} In Tab.~\ref{tab:tau}, we provide an analysis on $\tau$ defined in Eqn~\ref{eqn:edge}. The results indicate that the effects of variations in $\tau$ value are not significant and that a suitable threshold is $\tau$ = 0.2 for image input and $\tau$ = 0.3 for video input.

\paragraph*{Backbones}

In Tab.~\ref{tab:backbone}, we provide an ablation study with different transformer backbones. The \textit{``-S''} and \textit{``-B''} are ViT small~\cite{dosovitskiy2020image, caron2021emerging} and ViT base~\cite{dosovitskiy2020image, caron2021emerging} architecture respectively. The \textit{``-16''} and \textit{``-8''} represents patch sizes 16 and 8 respectively. The \textit{``DeiT''} is pre-trained supervised transformer model. The \textit{``MoCoV3''}~\cite{chen2021empirical} and \textit{``MAE''}~\cite{he2022masked} are pre-trained self-supervised transformer model. We optimise $\tau$ for different backbones: $\tau$ is set to 0.3 for MoCov3 and MAE, while for DINO and Deit $\tau$ is set to 0.2. Several insights can be found: \textit{1)} \name~is not suitable for supervised transformer models, while self-supervised transformers provide more powerful features allowing completing the task with \name. \textit{2)} As LOST~\cite{simeoni2021localizing} relies on a heuristic seeds expansion strategy, the performance varies significantly using different backbones. While our approach is more robust. Moreover, as no training is required for \name, it might be a more straightforward evaluation for the self-supervised transformers.

\begin{table}[!t]
	\begin{center}
	\caption{\textbf{Analysis of different bi-partition methods.} We report CorLoc for unsupervised single object discovery on VOC07~\cite{pascal-voc-2007}, VOC12~\cite{pascal-voc-2012} and COCO20K~\cite{lin2014microsoft,vo2020toward} datasets.}   
	
    \begin{tabular}{cccc}
    \toprule Bi-partition &
    VOC07 & VOC12 & COCO20K \\
    \midrule
    Mean        & \bf 68.8 & \bf 72.1 & 58.8 \\
    Energy (Eqn~\ref{eqn:ncut_energy}) & 67.3 & 69.7 & -\\
    EM          & 63.0 & 65.7 & 59.3 \\
    K-means      & 67.5 &  69.2 &  \bf 61.6  \\
    \bottomrule
    \end{tabular}
	\label{tab:partition}
	\end{center}
\end{table} 

\begin{table}[!t]
\begin{center}
\caption{\textbf{Analysis of video input.} We report Jaccard index for video segmentation on DAVIS~\cite{perazzi2016benchmark}, FBMS~\cite{ochs2013segmentation} and SegTV2~\cite{li2013video} with using input. ``RGB + Flow'' refers to using both video RGB frame and RGB representation of optical flow as input to the vision transformer. ``Mean\_Flow'' indicates using mean of flow instead of flow features extracted from vision transformer''. ``RGB'' and ``Flow'' present using only either RGB frames or optical flow as input. ``CRF'' indicates whether edge refinement step using CRF~\cite{krahenbuhl2011efficient} is computed.} %
\resizebox{\columnwidth}{!}{
\begin{tabular}{ccccc}
\toprule
Input   & CRF & DAVIS~\cite{perazzi2016benchmark}  & FBMS~\cite{ochs2013segmentation} & SegTV2~\cite{li2013video} \\
\midrule
\multicolumn{5}{c}{Graph per frame} \\
\midrule
RGB    & \cmark & 62.4    & 67.2 &   61.0          \\
Flow   & \cmark & 64.1  & 52.8 &    53.7         \\
\bf RGB + Flow & \cmark & 76.4  & 63.2 & \bf 64.4   \\
\midrule
\multicolumn{5}{c}{Graph per video} \\
\midrule
RGB    & & 51.8  & 58.4                &  59.3               \\
Flow&     & 49.9  & 48.3               & 46.7               \\
\bf RGB + Flow & & 64.3  &  60.2 &   59.6     \\
\midrule
RGB     & \cmark & 62.2  & \bf 67.5                &  63.7               \\
Flow   & \cmark & 63.1   & 50.2                & 50.2               \\
RGB + Mean\_Flow   & \cmark & 37.5  & 23.3  &    15.7 \\
\bf RGB + Flow   & \cmark & \bf 76.7  &  66.6 &  61.6     \\
\bottomrule
\end{tabular}}
\label{tab:video_ablation}
\end{center}
\end{table}

\begin{figure}[!t]
\centering
\begin{tabular}{c@{\hskip 3pt}c@{\hskip 3pt}c}

        \includegraphics[width=0.15\textwidth, height=0.13\textwidth]{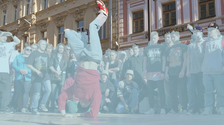} &
        \includegraphics[width=0.15\textwidth, height=0.13\textwidth]{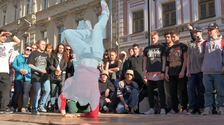}&
        \includegraphics[width=0.15\textwidth, height=0.13\textwidth]{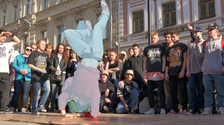} \\	
        
         \includegraphics[width=0.15\textwidth, height=0.13\textwidth]{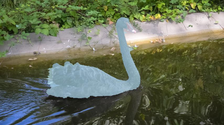} &
        \includegraphics[width=0.15\textwidth, height=0.13\textwidth]{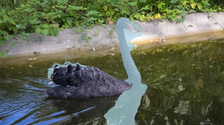}&
        \includegraphics[width=0.15\textwidth, height=0.13\textwidth]{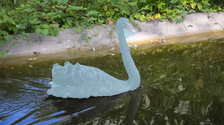} \\

        (a) RGB & (b) Flow & (c) RGB + Flow \\ 
\end{tabular}
\caption{\textbf{Visualization on DAVIS~\cite{perazzi2016benchmark} using different inputs.} We show segmentation results with RGB, Flow and RGB + Flow  in (a), (b) and (c) respectively.}
\label{fig: diff_inputs}
\end{figure}

\begin{figure}[!t]
\centering
\begin{tabular}{c@{\hskip 1.3pt}c@{\hskip 1.3pt}c@{\hskip 1.3pt}c}
    \rotatebox{90}{~~~~~\makecell{DAVIS}}
    \includegraphics[width=0.14\textwidth, height=0.12\textwidth]{rebuttal_horse06.png} &
    \includegraphics[width=0.14\textwidth, height=0.12\textwidth]{rebuttal_horse_07.png} &
    \includegraphics[width=0.14\textwidth, height=0.12\textwidth]{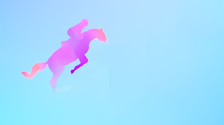} \\ 
    \rotatebox{90}{~~~~~\makecell{FBMS}}
     \includegraphics[width=0.14\textwidth, height=0.12\textwidth]{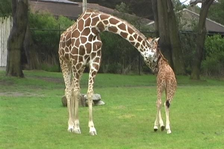} &
    \includegraphics[width=0.14\textwidth, height=0.12\textwidth]{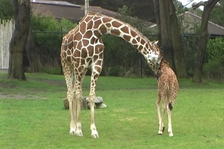} &
    \includegraphics[width=0.14\textwidth, height=0.12\textwidth]{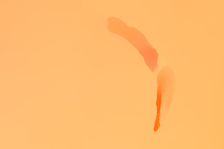}  \\ 
    \rotatebox{90}{~~~~~\makecell{SegTV2}}
    \includegraphics[width=0.14\textwidth, height=0.12\textwidth]{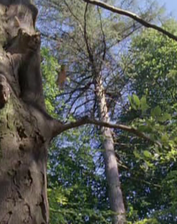} &
    \includegraphics[width=0.14\textwidth, height=0.12\textwidth]{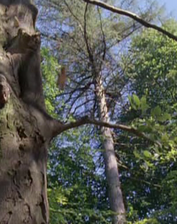} & 
    \includegraphics[width=0.14\textwidth, height=0.12\textwidth]{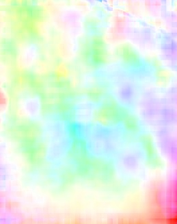}   \\ 
    (a) Frame t & (b) Next frame & (c) Flow\\
\end{tabular}
\caption{\textbf{Visual results of optical flow in DAVIS, FBMS, SegTV2}. In (a), we show Frame t. (b) is the following frame (t + 3 for FBMS and t + 1 for SegTV2) used to compute the optical flow. (c) is the optical flow using RAFT~\cite{teed2020raft}. Note that optical flow in the DAVIS dataset has high quality, whereas the flow in the FBMS and SegTV2, can sometimes fail, as in the cases shown in rows 2 and 3.} 
\label{fig:flow}
\end{figure}

\paragraph*{Bi-partition strategies}

In Tab.~\ref{tab:partition}, we study different strategies to separate the nodes in  into two groups using the second smallest eigenvector. We consider three natural methods: mean value (Mean), Expectation-Maximisation (EM), K-means clustering (K-means). We have also tried to search for the splitting point based on the best Ncut energy (Eqn~\ref{eqn:ncut_energy}). Note this approach is computationally expensive due to the quadratic complexity. The result suggests that the simple mean value as the splitting point performs well for most cases.

\paragraph*{Video input} 
We also study the impact of using RGB or optical Flow for video segmentation.  Quantitative results are presented in Tab.~\ref{tab:video_ablation}. We can see constructing graph on the entire video is better than constructing the graph per frame. We show an anaylsis by using the mean of optical flow feature and use it directly without feeding into transformer. The results illustrate that constructing the graph with RGB and RGB representation of the Flow together can significantly improve the performances over DAVIS~\cite{perazzi2016benchmark}. On FBMS~\cite{ochs2013segmentation} and SegTV2~\cite{li2013video}, due to the low quality of optical flow, the motion of salient objects are not detected in the optical flow as can be seen from the fact that they are not visible in the RGB visualisation. Some failure cases are shown in Fig.~\ref{fig:flow}. This failure of optical flow to detect slow motion impedes the inference process for augmenting appearance with optical flow features. The low quality of optical flow can be attributed to three factors: 1) small motion between two frames; 2) low quality of raw image, for instance several examples in SegTV2, such as birdfall; 3) the absence of fine-tuning for the pre-trained optical flow model on these three datasets.
Using both RGB appearance and flow lead to a slight improvement before  edge refinement, but slightly worse results  after edge refinement compared to using only RGB appearance. Some qualitative results are illustrated in Fig.~\ref{fig: diff_inputs}. We can see how RGB frame and optical flow are complementary to each other: in the first row, the target moving person shares semantically similar features to other audiences and using only RGB frames would produce a mask cover all the persons; in the second row, the flow also has non-negligible values on the surface of the river, thus using only flow leads to worse performance.

\begin{table}[!t]
\begin{center}
\caption{\textbf{Analysis of video graph.} We report Jaccard index~($\J$) for video segmentation on DAVIS~\cite{perazzi2016benchmark} with different video graphs. ``single frame'' represent creating the graph for each frame separately. }   
\resizebox{\columnwidth}{!}{
\begin{tabular}{ccc}
\toprule
Nodes         & Edges & DAVIS~($\J$) \\
\midrule
Video        & $\min (\simi(\feat^I_i, \feat^I_j), \simi(\feat^F_i, \feat^F_j))$              & 73.7      \\
Video        & $\max (\simi(\feat^I_i, \feat^I_j), \simi(\feat^F_i, \feat^F_j))$                 & 71.1      \\
Video        & $\frac{\simi(\feat^I_i, \feat^I_j)+\simi(\feat^F_i, \feat^F_j)}{2}$           & \bf 76.7         \\
Single Frame & $\frac{\simi(\feat^I_i, \feat^I_j)+\simi(\feat^F_i, \feat^F_j)}{2}$            & 76.4       \\
\bottomrule
\end{tabular}
}
\label{tab: graph_building}
\end{center}
\end{table}

\paragraph*{Video graph} 
In Tab.~\ref{tab: graph_building}, we provide an analysis for different ways to construct graphs for video. For edges, we also consider the minimum and maximum values between the flow and RGB similarities. For nodes, a natural baseline is to build a graph for each single frame. We can see that the optimal choice is to use the average value of the flow and RGB similarities (Eqn.~\ref{eqn:edge}) and build a graph for an entire video.

\begin{figure}[ht!]
\centering
\begin{tabular}{c@{\hskip 3pt}c@{\hskip 3pt}c@{\hskip 3pt}c}

        \includegraphics[width=0.11\textwidth, height=0.11\textwidth]{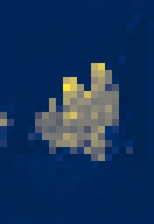} &
		\includegraphics[width=0.11\textwidth, height=0.11\textwidth]{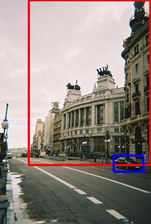} &
		\includegraphics[width=0.11\textwidth, height=0.11\textwidth]{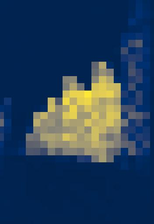} &
		\includegraphics[width=0.11\textwidth, height=0.11\textwidth]{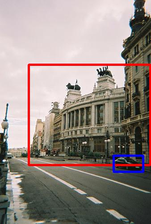}\\	
		
		\includegraphics[width=0.11\textwidth, height=0.11\textwidth]{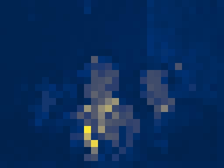} &
		\includegraphics[width=0.11\textwidth, height=0.11\textwidth]{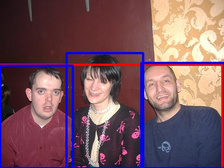} &
		\includegraphics[width=0.11\textwidth, height=0.11\textwidth]{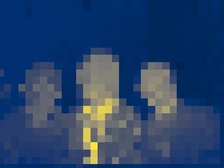} &
		\includegraphics[width=0.11\textwidth, height=0.11\textwidth]{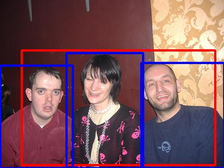}\\
		
		\includegraphics[width=0.11\textwidth, height=0.11\textwidth]{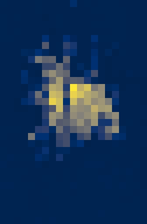} &
		\includegraphics[width=0.11\textwidth, height=0.11\textwidth]{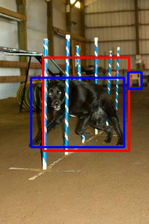} &
		\includegraphics[width=0.11\textwidth, height=0.11\textwidth]{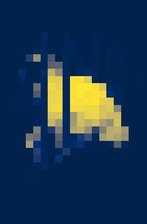} &
		\includegraphics[width=0.11\textwidth, height=0.11\textwidth]{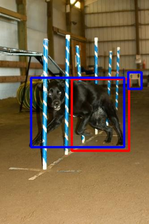}\\
        \makecell{(a) LOST \\ Inverse Attn.} & \makecell{(b) LOST \\Detection} & \makecell{(c) Our Eigen \\ Attention} & \makecell{(d) Our \\Detection} \\ 
\end{tabular}

\caption{\textbf{Failure cases on VOC12 (1st and 2nd row) and COCO (3rd row).} LOST~\cite{simeoni2021localizing}  mainly relies on the map of inverse degrees (a) to perform detection (b). For our approach, we illustrate the eigenvector in (c) and our detection in (d). \textcolor{blue}{Blue} and \textcolor{red}{Red} bounding boxes indicate the ground-truth and the predicted bounding boxes respectively.}
\label{fig:failure}
\end{figure}

\section{Discussion}
\begin{figure}[!ht]
    \centering
    \begin{tabular}{c@{\hskip 3pt}c@{\hskip 3pt}c@{\hskip 3pt}c}
        \includegraphics[width=0.11\textwidth, height=0.11\textwidth]{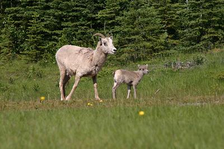} &
        \includegraphics[width=0.11\textwidth, height=0.11\textwidth]{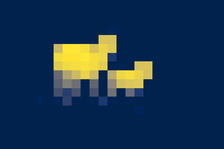} &
        \includegraphics[width=0.11\textwidth, height=0.11\textwidth]{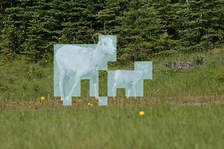} &
        \includegraphics[width=0.11\textwidth, height=0.11\textwidth]{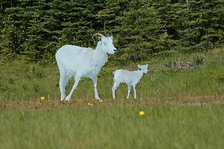} \\ 
        
        \includegraphics[width=0.11\textwidth, height=0.11\textwidth]{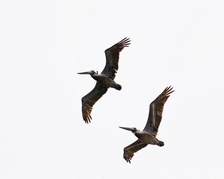} &
         \includegraphics[width=0.11\textwidth, height=0.11\textwidth]{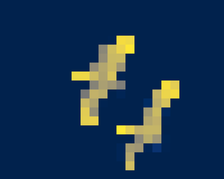} &
          \includegraphics[width=0.11\textwidth, height=0.11\textwidth]{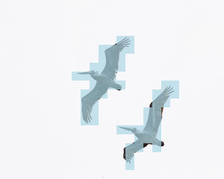} &
        \includegraphics[width=0.11\textwidth, height=0.11\textwidth]{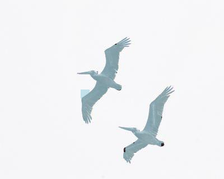} \\
        \makecell{(a) Raw \\ image} & \makecell{(b) Eigen \\ Attention} & \makecell{(c) Coarse \\ mask} & \makecell{(d) Fine \\ mask} \\
    \end{tabular}
    \caption{\textbf{Visual results of multiple objects in images}. In (a), we show the original image. TokenCut eigen Attention is illustrated in (b). (c) is TokenCut coarse mask segmentation results before selecting the largest connected component. (d) is TokenCut fine mask using bilateral solver.} 
    \label{fig: multiobject}
\end{figure}

\begin{figure}[ht!]
    \centering
    \begin{tabular}{c@{\hskip 1.3pt}c@{\hskip 1.3pt}c}
        \includegraphics[width=0.12\textwidth, height=0.1\textwidth]{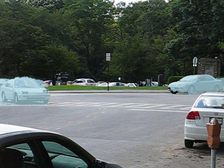} &
        \includegraphics[width=0.12\textwidth, height=0.1\textwidth]{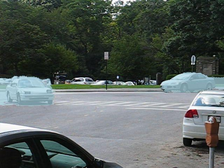} &
        \includegraphics[width=0.12\textwidth, height=0.1\textwidth]{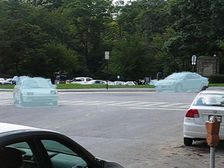}\\
        \includegraphics[width=0.12\textwidth, height=0.1\textwidth]{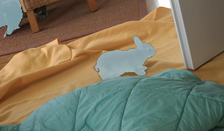} &
        \includegraphics[width=0.12\textwidth, height=0.1\textwidth]{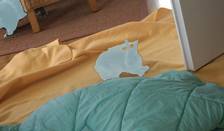} &
        \includegraphics[width=0.12\textwidth, height=0.1\textwidth]{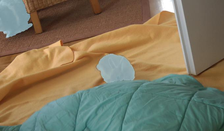} \\
        (a) Frame t & (b) Frame t+5 & (c) Frame t+10\\
    \end{tabular}
    \caption{\textbf{Visual results of multiple objects moving.} We visualize the case when multiple object moving from different direction, TokenCut is also capable of segmenting the moving objects.}
    \label{fig: moving}
\end{figure}

\paragraph*{Multi-Object Segmentation}
In the context of the Unsupervised Single Object Discovery task, the primary objective is to identify the most salient object within a given image. Consequently, we only choose the largest connected component in our approach. However,  \name  can identify more than one connected component in the second smallest eigenvector when multiple objects are present in the images. To illustrate this capability, we have included two examples in Fig~\ref{fig: multiobject}. In Fig~\ref{fig: moving}, we provide examples when multiple objects are moving from different directions. These results illustrate the robustness of our method.

\paragraph*{Limitations} 
Despite the good performance of the {\name} proposal, it has several limitations. We show several failure cases in Fig.~\ref{fig:failure}:
i) As seen in the 1st row, {\name} focuses on the largest salient part in the image, which may not be the desired object.
ii) Similar to 
LOST~\cite{simeoni2021localizing}, {\name} assumes that a single salient object occupies the foreground. If multiple overlapping objects are present in an image, both LOST and our approach would fail to detect one of the object, as displayed in the 2nd row. 
iii) For object detection, neither LOST nor\name~can handle occlusion properly, as shown in the 3rd row.

\section{Conclusion}

This paper describes {\name}, an unified and effective approach for both image and video object segmentation without the need for supervised learning. \name~uses features from self-supervised transformers to constructs a graph where nodes are patches and edges represent similarities between patches. For videos, optical flow is incorporated to determine moving objects. We  show that salient objects can be directly detected and delimited using the Normalized Cut algorithm. We evaluated this approach on unsupervised single object discovery, unsupervised saliency detection, and unsupervised video object segmentation, demonstrating that  \name~can provide a significant improvement over previous approaches. Our results demonstrate that self-supervised transformers can provide a rich and general set of features that may likely be used for a variety of computer vision problems.

\section*{Acknowledgment}

This work has been partially supported by the  MIAI  Multidisciplinary AI Institute at the Univ. Grenoble Alpes (MIAI@Grenoble Alpes - ANR-19-P3IA-0003),  and by the EU H2020 ICT48 project Humane AI Net under contract EU \#952026.

\bibliographystyle{ieee_fullname}
\bibliography{IEEEabrv, IEEEexample}

\end{document}